\documentclass[sn-mathphys,Numbered]{sn-jnl}

\usepackage{graphicx}%
\usepackage{multirow}%
\usepackage{amsmath,amssymb,amsfonts}%
\usepackage{amsthm}%
\usepackage{mathrsfs}%
\usepackage[title]{appendix}%
\usepackage{xcolor}%
\usepackage{textcomp}%
\usepackage{manyfoot}%
\usepackage{booktabs}%
\usepackage{algorithm}%
\usepackage{algorithmicx}%
\usepackage{algpseudocode}%
\usepackage{listings}%
\usepackage{comment}

\theoremstyle{thmstyleone}%
\theoremstyle{thmstyletwo}%

\theoremstyle{thmstylethree}%
\newtheorem{definition}{Definition}%

\newtheorem{problem}{Problem}%

\begin{document}

\title[Article Title]{Model-free Motion Planning of Autonomous Agents for Complex Tasks in Partially Observable Environments}


\author*[1]{\fnm{Junchao} \sur{Li}}\email{junchao-li@uiowa.edu}
\equalcont{These authors contributed equally to this work.}

\author[2]{\fnm{Mingyu} \sur{Cai}}\email{mic221@lehigh.edu}
\equalcont{These authors contributed equally to this work.}

\author[3]{\fnm{Zhen} \sur{Kan}}\email{zkan@ustc.edu.cn}

\author[4]{\fnm{Shaoping} \sur{Xiao}}\email{shaoping-xiao@uiowa.edu}
\equalcont{These authors contributed equally to this work.}

\affil*[1]{\orgdiv{Department of Mechanical Engineering, Iowa Technology of Institute}, \orgname{The University of Iowa}, \orgaddress{\street{Seamans Center, 3131}, \city{Iowa City}, \postcode{52242}, \state{IA}, \country{USA}}}

\affil[2]{\orgdiv{Department of Mechanical Engineering}, \orgname{Lehigh University}, \orgaddress{\street{19 Memorial Dr W}, \city{Bethlehem}, \postcode{18015}, \state{PA}, \country{USA}}}

\affil[3]{\orgdiv{Department of Automation}, \orgname{University of Science and Technology of China}, \orgaddress{\street{No. 443 Huangshan Road, Shushan District}, \city{Hefei}, \postcode{230022}, \state{Anhui}, \country{China}}}

\affil[4]{\orgdiv{Department of Mechanical Engineering, Iowa Technology of Institute}, \orgname{The University of Iowa}, \orgaddress{\street{Seamans Center, 3131}, \city{Iowa City}, \postcode{52242}, \state{IA}, \country{USA}}}


\abstract{Motion planning of autonomous agents in partially known environments with incomplete information is a challenging problem, particularly for complex tasks. This paper proposes a model-free reinforcement learning approach to address this problem. We formulate motion planning as a probabilistic-labeled partially observable Markov decision process (PL-POMDP) problem and use linear temporal logic (LTL) to express the complex task. The LTL formula is then converted to a limit-deterministic generalized Büchi automaton (LDGBA). The problem is redefined as finding an optimal policy on the product of PL-POMDP with LDGBA based on model-checking techniques to satisfy the complex task. We implement deep Q learning with long short-term memory (LSTM) to process the observation history and task recognition. Our contributions include the proposed method, the utilization of LTL and LDGBA, and the LSTM-enhanced deep Q learning. We demonstrate the applicability of the proposed method by conducting simulations in various environments, including grid worlds, a virtual office, and a multi-agent warehouse. The simulation results demonstrate that our proposed method effectively addresses environment, action, and observation uncertainties. This indicates its potential for real-world applications, including the control of unmanned aerial vehicles (UAVs).}

\keywords{motion planning, partially observable environments, complex tasks, linear temporal logic, reinforcement learning, recurrent neural networks}



\maketitle

\section{Introduction}\label{sec1}

The partially observable Markov decision process (POMDP) \cite{Kurniawati2022} provides a mathematical framework to model decision-making problems, including motion planning of autonomous agents, e.g., unmanned aerial vehicles (UAVs). It differs from Markov decision processes (MDPs) \cite{Kaufman1961} that have been widely applied in robotics and autonomous systems, assuming the environment is fully observable. POMDPs are more realistic for real-world applications in which the agent (e.g., the robot) may lack enough information from perception and cannot completely identify the state of the environment. On the other hand, simple go-to-goal motion planning tasks have been extensively studied by conventional pathfinding techniques \cite{perez2012} and reinforcement learning (RL) methods \cite{sutton2018}. However, complex tasks like surveillance missions are more relevant to real-world applications. Therefore, it is challenging for the agent to learn how to plan its motions to accomplish complex tasks in partially observable environments, especially considering environment and action uncertainties. 

Over the past decade, researchers have attempted to solve POMDP problems (with simple go-to-goal tasks) by using various model-based reinforcement learning (RL) algorithms. Many modern solvers can handle large spatial domains with thousands of states \cite{Shani2013}. A commonly-used approach includes point-based value iteration (PBVI) \cite{Pineau2003, Kurniawati2009} methods, consisting of model-based algorithms to approximately solve the POMDP problems by computing a value function over a finite subset of the belief space. It shall be noted that a belief state represents a probability distribution of the states where the agent can be. After each transition, the belief state needs to be updated by the transition and observation probabilities via the Bayesian approach \cite{Icarte2019}. Indeed, such model-based approaches transform a POMDP problem into an equivalent MDP problem with the corresponding belief state space.

Another solution to POMDP problems is the model-free RL approach, in which the agent doesn't know transition and observation probabilities. Consequently, the policy maps a sequence of observations (i.e., the observation history) to the selected action. Mnih \textit{et al}.\cite{Mnih2013, Mnih2015} first introduced a deep Q-Learning (i.e., DQN) and tested it on several Atari 2600 games, in which the Q networks were trained to reach human-level performances. Particularly, their Q networks took the last four frames (in grayscale) as the input and utilized a convolutional neural network (CNN) \cite{Krizhevsky2017} to extract the image features for a fully-connected neural network to predict Q values (i.e., the state-action values). However, since the environment was modeled as MDPs in their work, this approach bypassed the issue of partial observability, and the last four frames allowed the agent to access limited past experiences.

Based on the above-mentioned works, Hausknecht and Stone \cite{Hausknecht2015} proposed adding a recurrent neural network (RNN) to the Q network architecture. They modeled the Atari 2600 games as POMDP problems and proposed a so-called deep recurrent Q-networks (DRQNs) by replacing the first post-convolutional fully-connected layer of Q networks with a long short-term memory (LSTM) \cite{Hochreiter1997}, which took a single image at each time step as the input feature. Hence, this approach was able to integrate the entire observation history instead of utilizing the observation sequences with a fixed length (e.g., four frames of images as the input in \cite{Mnih2013}). In addition, Foerster \textit{et al}.\cite{Foerster2016} extended the DRQN to handle multi-agent RL problems in partially observable environments by proposing a deep distributed recurrent Q-networks (DDRQN), where the last action was fed as the input to the Q network. Zhu \textit{et al}.\cite{Zhu2017} developed a new network architecture called action-specific deep recurrent Q-network (ADRQN), in which an LSTM layer processed the action history and associated observations for the Q value computing. Some other similar works, including \cite{Heess2015, Meng2021}, implemented recurrent neural networks (RNNs) for the control policy on POMDP problems with continuous state spaces.

It shall be noted that the works on POMDP problems mentioned above consider simple go-to-goal missions only. However, complex tasks have been included in MDP problems via formal languages \cite{Baier2008}, such as linear temporal logic (LTL). Generally, user-defined high-level specifications can be expressed as an LTL formula, which is then converted to an $\omega$-automaton over infinite words with a B\"{u}chi or a Rabin acceptance condition \cite{Kretinsky2018}. Consequently, robotic motion planning problems can be solved via control synthesis for a product of MDP and automaton. Recently, this formal approach was employed to verify the task objectives when solving POMDP problems with certain temporal logic constraints. Chatterjee \textit{et al.} \cite{Chatterjee2015} studied the undecidability of the qualitative model checking in an infinite-horizon POMDP. Their approach relied on exploring the entire belief space and was most suitable to the problems with small state spaces. They also concluded that it might be unable to acquire the optimal policy, ensuring the maximum satisfaction probability of the specific logic formula in POMDPs.

Other works like \cite{Sharan2014, Bouton2020, Ahmadi2020} proposed solving such problems by converting LTL specifications to a deterministic Rabin automaton (DRA), then constructing a product of POMDP and DRA. Specifically, Sharan \textit{et al.} \cite{Sharan2014} and Ahmadi \textit{et al.} \cite{Ahmadi2020} employed finite state controllers (FSCs) to limit the policy search via the value iteration method. Also, Bouton \textit{et al.} \cite{Bouton2020} utilized the approximate POMDP solver, SARSOP \cite{Kurniawati2009}, to search for an optimal policy on the finite belief state space of the product POMDP. However, those approaches are model-based RL methods, which require the agent to know the transition and observation probabilities. Such a requirement limits the applications in unknown environments. It shall be noted that the most related work that employs RNNs and model-checking for POMDP problems satisfying the temporal logic constraints was carried out by Carr \textit{et al.} \cite{Carr2019, Carr2020}. Their proposed method focused on constructing a policy-based procedure to iteratively improve the current policy by implementing RNN. However, their approach constructed an underlying MDP to map the original POMDP and utilized a model-based RL solver, SARSOP \cite{Kurniawati2009}, to approximate the value function. Hence, as they stated, it was still a model-based approach.

On the other hand,  It has been shown in \cite{Hahn2019} that a limit-deterministic B\"{u}chi automaton (LDBA) has more advantages than a DRA in model-free and model-based RL learning when solving MDP problems. Hasanbeig \textit{et al.} \cite{Hasanbeig2019} stated that converting the LTL specification into an LDBA might result in a smaller automaton state space than a DRA. Furthermore, unlike LDBAs, a limit-deterministic generalized B\"{u}chi automaton (LDGBA) has multiple accepting sets and enables the agent can visit all accepting sets infinitely often. In addition, the conversion of LDGBA avoids the sparsity of reward caused by LDBA, which may slow down the convergence in RL \cite{Oura2020}. 

In this paper, we model the interactions between the agent and its partially observable surroundings as a probabilistic-labeled POMDP (PL-POMDP). Introducing probabilistic labels in a POMDP enables us to consider both static and dynamic events in the environment. One of the contributions of this work is converting the LTL formula to an LDGBA, which represents LTL specifications for a complex task in the considered POMDP problem. This has not been reported in previous works, according to the authors' best knowledge. After generating a product of PL-POMDP and LDGBA, the original problem of finding a policy that satisfies LTL specifications in a PL-POMDP can be reformulated as finding an optimal policy to maximize the collected reward on the corresponding product POMDP. Another contribution is proposing a model-free RL approach to learn an optimal policy on the product POMDP. We implement an RNN into Q network architectures to process the information that the agent acquires: the observation history and the task recognition. The latter depends on whether the agent fully recognizes the LTL-induced automaton. Specifically, either the history of automation states or the history of the state labels is utilized, respectively, 

This paper is organized as follows: Section 2 reviews the PL-POMDP definition and introduces deep Q-learning with RNN to solve a simple go-to-goal POMDP problem. Section 3 presents LTL, LDGBA, and product PL-POMDP. Then, the problem of PL-POMDP with LTL specifications is reformulated. Section 4 proposes the model-free approaches to solve product PL-POMDP problems and provides detailed algorithms. Finally, experiments and results are included in Section 5, followed by the discussions and conclusions, including future works.

\section{Background}\label{sec2}

In this section, we first define the probabilistic-labeled POMDP (PL-POMDP) and then explain using DQN, a model-free RL method, to solve a POMDP problem with simple go-to-goal tasks. Finally, a simulation example is conducted for the demonstration. Our focus in this paper is extending the DQN for the POMDP problem with complex tasks, and the developed methodology will be detailed in Section~\ref{sec4}. 

\subsection{PL-POMDP}\label{sec2a}

The POMDP is usually adopted as a mathematical description of problems in which the agent cannot fully observe and completely identify its surroundings. We employ PL-POMDP with the consideration of static and dynamic events. 

\begin{definition}[PL-POMDP] \label{def:POMDP} 
Considering the transition and observation uncertainties and the probabilistic labels of states, we can denote a PL-POMDP by a tuple $\mathcal{P}=\left(S, A, T, s_0, R, O, \Omega, \Pi, L, P_L\right)$, which consists of:
\begin{itemize}
\item A finite set of states, $S = \{s_1, . . . , s_n\} $. 
\item A finite set of actions, $A = \{a_1, . . . , a_m\}$. Particularly, $A(s)$ is the set of actions available for the agent at the current state $s$.
\item A transition probability function, $T: S\times A\times S\rightarrow \left[0,1\right]$ when the agent moves from the current state $s \in S$ to the next state $s' \in S$ after executing an action $a \in A(s)$. There exists $\sum_{s'\in S} T(s,a,s') = 1$. 
\item An initial state, $s_0 \in S$.
\item A reward function, $R: S \times A \times S \rightarrow R$. It shall be noted that reward function sometimes can be defined as $R(s')$, $R(s, a)$ or $R(a, s')$.
\item A finite set of observations, $O = \{o_1, . . . , o_k\}$. At the current state $s$, $O(s)$ consists of possible observations the agent can perceive.
\item An observation probability function, $\Omega: S \times A \times O \rightarrow \left[0,1\right]$, represents the probability that the agent can perceive observation $o$ at state $s' \in S$ after executing action $a \in A(s)$. This function follows $\sum_{o\in O(s')} \Omega(s',a,o) = 1$.
\item A set of atomic propositions, $\Pi$.
\item A labeling function, $L: S\rightarrow 2^{\Pi}$, outputs a set of all possible labels at state $s$. $2^{\Pi}$ is the power set of $\Pi$.
\item A labeling probability function, $P_L(s,l)$, represents the probability of a label $l \in L(s)$ associated with a state $s \in S$. It satisfies $\sum_{l \in L(s)}P_L(s,l) = 1, \forall s \in S$.
\end{itemize}
\end{definition}

When the agent interacts with its environment, it has a (transition) probability $T(s, a, s')$ to move from state $s \in S$ to state $s’ \in S$ after choosing and executing an action $a \in A(s)$. At the next state $s'$, the agent has an (observation) probability $\Omega(s', a, o)$ to perceive an observation $o \in O(s')$. In addition, the agent receives a reward as feedback from the environment based on the reward function $R(s, a, s')$.

The states' labels are atomic propositions representing event occurrences at specific states. In addition, the labeling probability function can characterize a static event $l$ at state $s$ if $L(s)=\{l\}$ and $P_L(s,l) = 1$ or a dynamic event otherwise. The labels indicate goal states to handle simple go-to-goal tasks. When considering a complex task, we can use formal language to formulate the task regarding labels and then convert the formula to a finite state automaton. The labels are also input symbols for the induced automaton. Therefore, the accomplishment of the complex task can be validated via model checking during the motion planning. We will provide the details of such an approach in Section~\ref{sec3}.  

\subsection{DQN for POMDP problems}\label{sec2b}

To solve a POMDP problem with unknown transition probability, DQN is employed to map a sequence of observations to Q values for action selection. We consider observation histories up to $j$ previous time steps, so the sequence of observations is denoted as $\mathbf{o}_t=(o_{t-j}, o_{t-j+1}, o_{t-j+2}, ..., o_t)$ with a length of $j+1$. Consequently, the policy is a function of observation sequence. The agent's objective in deciding a course of action is maximizing the expected return as below, representing the total discounted rewards the agent can collect from the current time under a policy $\xi$.
\begin{equation}\label{eq:expReturn} 
U^{\xi} (s_t) = \mathbb{E}^{\xi} \left[\sum_{\tau=0}^{\infty} \gamma^{\tau}  R(s_{t+\tau}, a, s_{t+\tau+1}) \Big \vert s_t \right]
\end{equation} 
where $s_{t}$ is the agent's state at time $t$, and $\gamma \in [0,1]$ is the discount factor to balance the importance between immediate and future rewards. Then, the optimal policy can be found as $\boldsymbol{\xi}^{*} (\mathbf{o}_t)={\mathrm{argmax}}_{\boldsymbol{\xi}}U^{\boldsymbol{\xi}}\left( s_0 \right)$. 

\begin{figure}
\centering
\includegraphics[width=120mm]{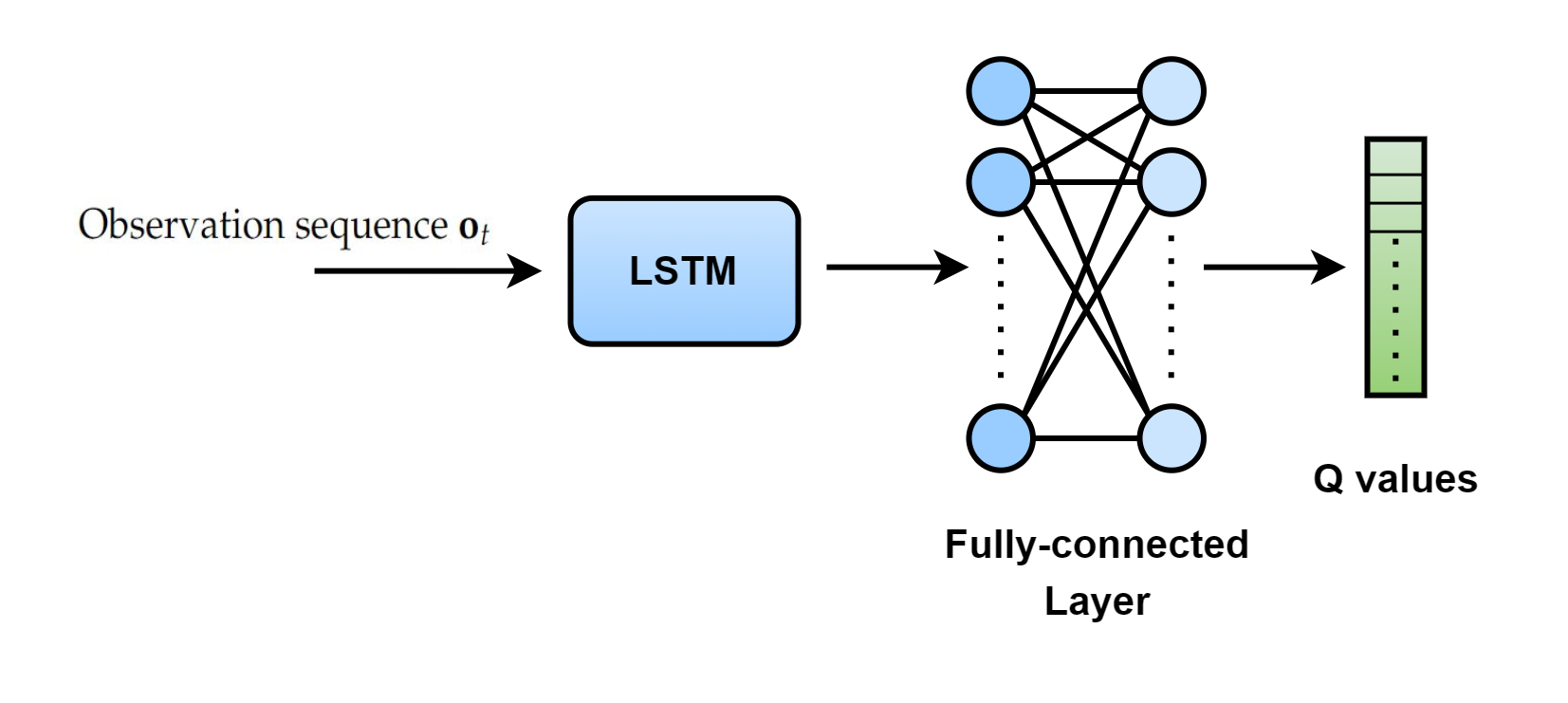}
\caption{Q network architecture.}\centering\label{fig:DRQN architecture 1}
\end{figure}

We implement an LSTM to process the observation sequence in the Q network architecture as shown in Figure~\ref{fig:DRQN architecture 1}. LSTM is one type of RNN that can model temporal dependencies between observations by introducing feedback loops in the network architecture. After processing the observation at each time step, the output is fed back into the network as input (together with the following observation) for the next time step. This allows the network to capture information about the order and timing of observations in the sequence. One-dimensional convolutional neural networks (CNNs) and deep neural networks (DNNs) can also be used to process time-series data. However, they don’t explicitly model the temporal dependencies between observations, and DNNs cannot process input sequences of variable length. In the next subsection, we use a simple go-to-goal simulation example to demonstrate the advantages of RNN in Q networks over CNN and DNN. 

DQN usually has two Q networks: an evaluation Q-network $Q_E(\mathbf{o}_t, a_t; \theta_E)$ and a target Q-network $Q_T(\mathbf{o}_t, a_t; \theta_T)$, where $\theta_E$ and $\theta_T$ are the network weights, respectively. The evaluation Q-network $Q_E$ is usually updated at each iteration by randomly selecting a batch of data samples from a so-called replay memory \cite{Lin1992} during the learning process. At the same time, the target Q-network $Q_T$ keeps fixed weights until copying from the evaluation Q-network $Q_E$ once in a while, i.e., $\theta_T = \theta_E$. 

At each step (e.g., time $t$) during the learning process, the target Q-network ($Q_T$) predicts Q values based on the sequence of observations $\mathbf{o}_{t}$ so that the agent can choose an action $a_t$ via the $\epsilon$-greedy technique \cite{sutton2018}. After executing the selected action and perceiving an observation $o_{t+1}$, a new sequence of observation $\mathbf{o}_{t+1} = (o_{t-j+1}, o_{t-j+2}, o_{t-j+3}, ..., o_{t+1})$ is generated. Consequently, an experience is formed as $e_t = (\mathbf{o}_t, a_t, r_t, \mathbf{o}_{t+1})$, where $r_t = R(s_t, a_t, s_{t+1})$. The experience is recorded as one data sample in the replay memory $D$, where the data samples are randomly selected to update the new Q values for Q network updating. The equation used to update the Q value associated with $a_t$ is defined below. 
\begin{equation}\label{eq:DRQN Q value}
Q_{new}(\mathbf{o}_t, a_t) = Q_E(\mathbf{o}_t,a_t; \theta_E)+\alpha \left[ r_t+\gamma \max_{a_{t+1}} Q_T(\mathbf{o}_{t+1},a_{t+1}; \theta_T)-Q_E(\mathbf{o}_t,a_t; \theta_E) \right]
\end{equation}
where $\alpha$ is the learning rate. 

\subsection{A go-to-goal example}\label{sec2c}

\begin{figure}
\centering
\includegraphics[width=50mm]{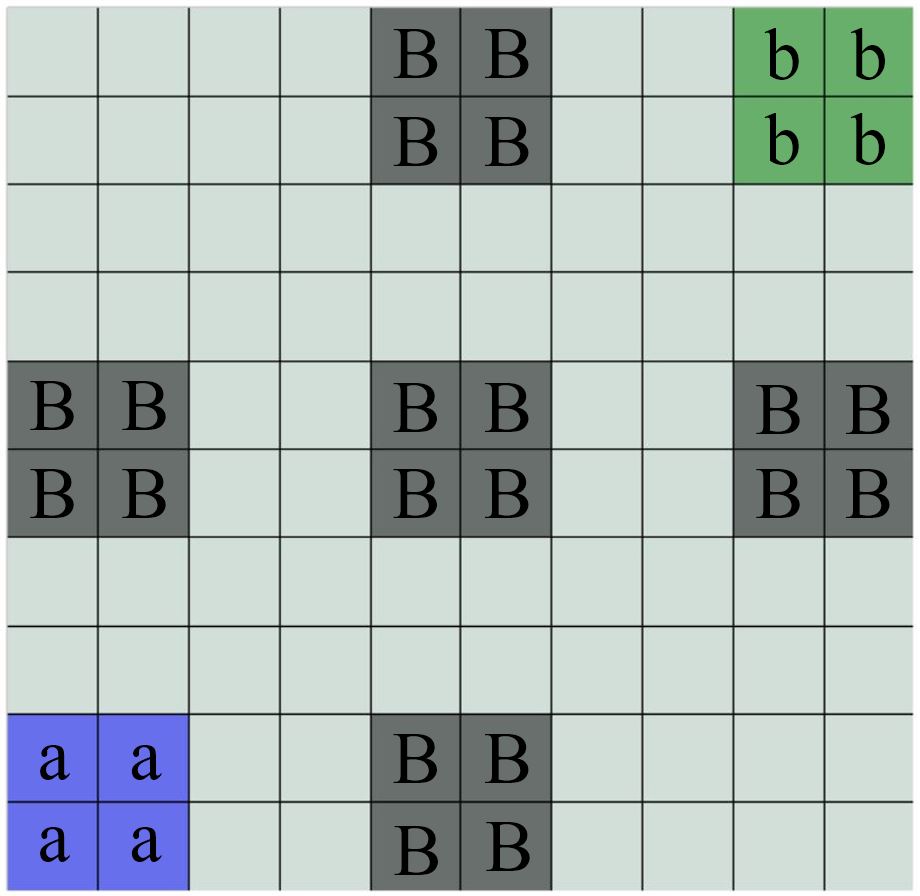}
\centering\caption{A 10 x 10 grid world}
\label{fig:grid-world_1}
\end{figure}

We take a simple go-to-goal example in a grid world to demonstrate DQN, i.e., a model-free RL method, to solve a POMDP problem. Figure \ref{fig:grid-world_1} illustrates a 10 × 10 grid world, in which the blue area labeled as `a' and the green area labeled as `b' are the initial and goal state, respectively. The block states, i.e., obstacles, are labeled with `B'. The agent, e.g., a mobile robot in the grid world, must move from the initial state to the goal state. It can take four actions at each state: $up$, $left$, $down$, and $right$. Due to the action uncertainty, the agent has a probability of 0.9 to take the selected action. Otherwise, it takes two side-way actions with equally weighted probabilities. In addition, the agent will remain in the current state if the next state is outside the grid world or the obstacles. After reaching the next state, the agent can observe this state with a probability of 0.9 and adjacent states with a total probability of 0.1 uniformly distributed. The discount factor $\gamma$ is set as 0.98.

It shall be noted that the observations are states in this grid world example. Therefore, the observation sequence is pre-processed by the technique of one-hot encoding since the row and column indices of a state are ordinal data. The generated 1-D vector as the input is fed into the LSTM layer. The extracted feature is then passed to two fully-connected layers with 16 neurons, respectively, to predict the corresponding Q values. The rectified linear unit (ReLU) is utilized as the activation function in the Q networks. The learning process for this problem includes 500 steps per episode for 1,000 episodes. Two Q networks, the evaluation network $Q_E$ and the target network $Q_T$, are randomly initialized. The training batch size for the evaluation network is 32, and the target network is updated by copying the weight coefficients of $Q_E$ every 50 steps.

\begin{figure}
\centering
\includegraphics[width=80mm]{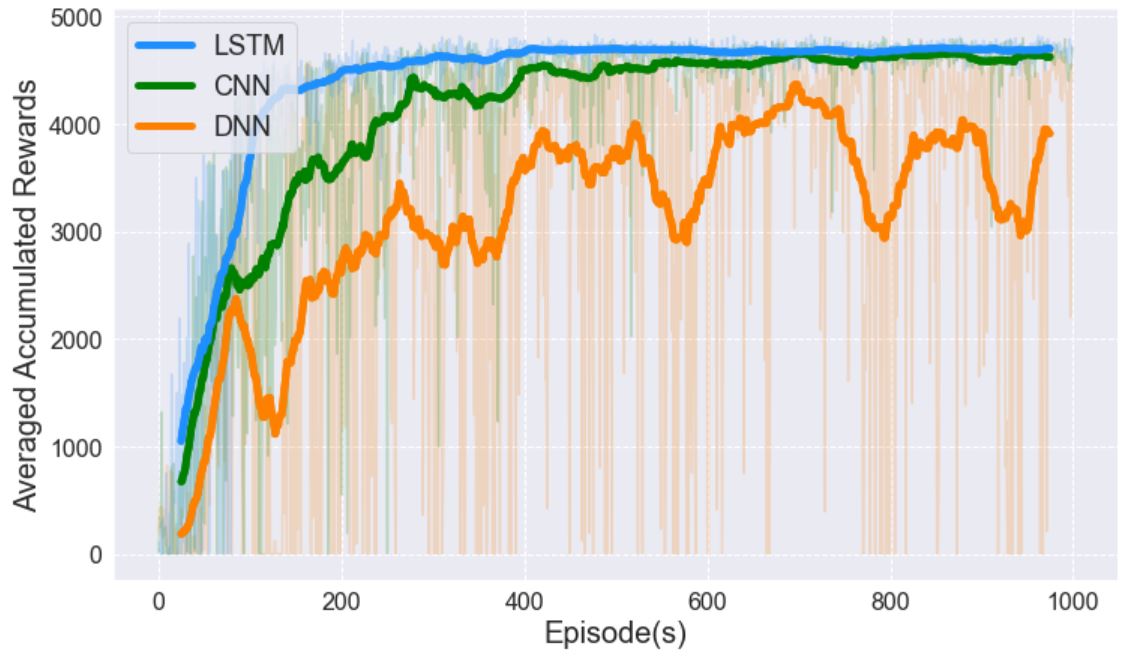}
\caption{The comparison of accumulated rewards by using Q networks with LSTM, CNN, and DNN.}\centering\label{fig:go_to_goal_plot}
\end{figure}

We also test two other Q network architectures by implementing CNN and deep neural network (DNN). CNN-based Q networks use a 2D convolutional layer (filter=6, kernel size=(3,3), strides=1) followed by another 2D convolutional layer (filter=12, kernel size=(2,2), strides=1) before the fully connected layers. The DNN uses the fully connected layers only. Figure \ref{fig:go_to_goal_plot} compares the averaged accumulated rewards collected by the agent every ten episodes (the trend lines represent the simple moving average (SMA) of rewards every 50 episodes). The results illustrate that Q networks with LSTM can achieve faster convergence and a higher accumulated reward.  

\begin{figure}
\centering
\includegraphics[width=50mm]{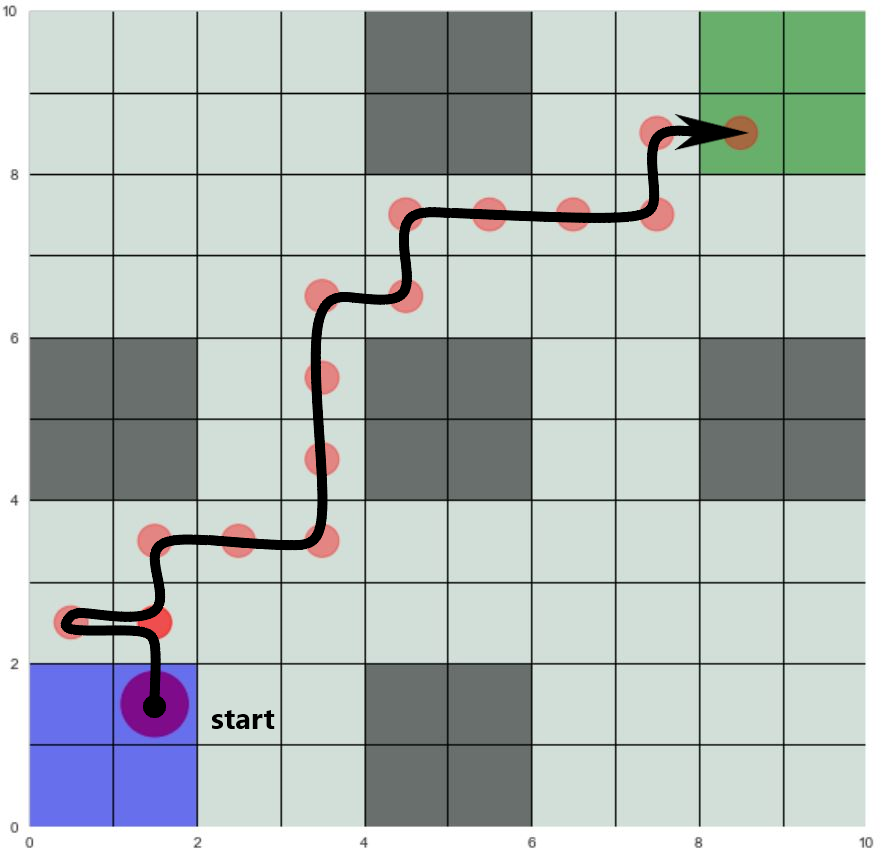}
\caption{A path generated from the derived policy.}\centering\label{fig:go_to_goal_path}
\end{figure}

An optimal policy can be derived from the Q networks after convergence. Figure~\ref{fig:go_to_goal_path} displays a path generated from the derived policy for the agent moving on the grid world to accomplish this simple go-to-goal task. The start state is marked as the large purple solid circle, and the reached states are marked as light red dots. The brighter red dot and the bend of the black route indicate this state has been visited more than once. It shall be noted that even if the policy converges and approaches the optimal, the generated path may not. This is because of observation and action uncertainties.

\section{Problem Definition}\label{sec3}

In this study, we introduce a framework for solving a PL-POMDP problem with LTL specifications by transforming the LTL formula into an LDGBA that represents the task variables and safety constraints of the POMDP and generating a product of POMDP and LDGBA. The problem of satisfying the given LTL constraints in a POMDP is equivalent to the problem of reaching (Büchi) accepting states in the product POMDP. 

\subsection{Linear temporal logic (LTL)}\label{sec3a}

Linear temporal logic \cite{Baier2008} is a logical formalism for linear-time properties, representing the relation between state labels in sequential executions. In addition to the Boolean connectors, LTL extends propositional logic by adding some temporal operators, including two basic ones $\bigcirc$ (pronounced ``next") and $\mathcal{U}$ (pronounced ``until"). This study assumes that $a \in \Pi$ is an atomic proposition, and $\phi, \phi_1$ and $\phi_2$ are single LTL formulas. Then, LTL formulas can be formed according to the following grammar \cite{Bozkurt2020}:  
\begin{equation}\label{eq:LTL}
        \phi   :=  \text{True} \mid a \mid \phi_1 \land \phi_2 \mid \lnot \phi \mid \bigcirc \phi \mid \phi_1 \mathcal{U} \phi_2 
\end{equation}
where negation (¬) and conjunction ($\wedge$) are the Boolean operators. Formula $\bigcirc \phi$ is true at the current time if $\phi$ is true the next time. In addition, Formula $\phi_1 \mathcal{U} \phi_2$ is true at the current time if $\phi_2$ is true for some future time and $\phi_1$ is true at all times until that future time. 

Other commonly-used temporal operators are $\diamondsuit$ (pronounced ``eventually") and $\square$ (pronounced ``always"). Formula $\diamondsuit \phi$ ensures that $\phi$ will be true eventually in the future, while $\square \phi$ is true from now on forever. They can be derived as follows: 
\begin{equation}
\arraycolsep=1.4pt
\begin{array}{lcl}
\text{eventually}: && \diamondsuit \phi \equiv \text{True }\mathcal{U} \phi\\
\text{always}: && \square \phi \equiv \lnot(\diamondsuit \lnot\phi) \\
\end{array} 
\end{equation} 
When using $\models$ to represent the satisfaction relationship, we can interpret the semantics of an LTL formula over words as below. A word is an infinite sequence $\boldsymbol{w}=w_{0}w_{1}\ldots$ with $w_{i}\in2^{\Pi}$ where $\Pi$ is a set of atomic propositions for all $i\geq0$.
\begin{equation}
\arraycolsep=1.4pt
\begin{array}{lcl}
\boldsymbol{w} \models \text{True}  \\
\boldsymbol{w} \models \alpha  & \Leftrightarrow & \alpha\in  L(\boldsymbol{w}[0])  \\
\boldsymbol{w} \models \phi_{1}\land\phi_{2} &  \Leftrightarrow & \boldsymbol{w} \models \phi_{1} \text{ and } \boldsymbol{w} \models \phi_{2}  \\
\boldsymbol{w} \models \lnot\phi  & \Leftrightarrow & \boldsymbol{w} \mid\neq\phi  \\
\boldsymbol{w} \models \bigcirc\phi  & \Leftrightarrow & \boldsymbol{w}[1:] \models\phi  \\
\boldsymbol{w} \models \phi_1 \mathcal{U} \phi_2  & \Leftrightarrow & \exists t \text{ s.t. }\boldsymbol{w}[t:]\models\phi_{2}, \forall t'\in [0,t),  \boldsymbol{w}[t':]\models\phi_{1}  \\
\end{array} 
\end{equation}

\subsection{Limit-deterministic generalized B\"uchi automaton (LDGBA)} \label{sec3b}

The previous subsection describes that a user-specified complex task can be formulated via LTL. Then, we can convert the LTL formula into an automaton, including LDGBA \cite{Sickert2016}, to evaluate task satisfaction via model checking \cite{Baier2008}.
\begin{definition}[LDGBA] \label{def:LDGBA} 
An LDGBA  $\mathcal{A}=\left(Q,\Sigma,\delta,q_{0},\mathcal{F}\right)$ consists of a finite set of states $Q$, a finite alphabet (i.e., a finite set of input symbols) $\Sigma=2^{\Pi}$ where $\Pi$ is a set of atomic propositions, a transition function $\delta\colon$ $Q\times\left(\Sigma\cup\left\{ \epsilon\right\} \right)\rightarrow2^{Q}$, an initial state $q_{0}\in Q$, and a set of accepting sets $\mathcal{F}=\left\{ \mathcal{F}_{1},\mathcal{F}_{2},\ldots,\mathcal{F}_{f}\right\} $ where $\mathcal{F}_{i} \subseteq Q$, $\forall i\in \left\{ 1,\ldots, f \right\} $. Furthermore, the state set $Q$ can be decomposed into deterministic and non-deterministic sets, i.e., $Q_{D}$ and  $Q_{N}$, respectively. They satisfy the following requirements. 
	\begin{itemize}
            \item $Q_{D}\cup Q_{N}=Q$ and $Q_{D}\cap Q_{N}=\emptyset$. 
		\item The state transitions in $Q_{D}$ are total, i.e., $\vert \delta\left(q,\alpha\right)\vert =1$. 
            \item The transitions in $Q_D$ are restricted within
		it, i.e., $\delta\left(q,\alpha\right)\subseteq Q_{D}$
		for every state $q\in Q_{D}$ and $\alpha\in\Sigma$. 
		\item The $\epsilon$-transitions do not take the input symbols and are only valid from $q \in Q_{N}$ to $q' \in Q_{D}$.
		\item The accepting sets, consisting of accepting states, are only defined in the deterministic set. In other words, $\mathcal{F}_{i}\subseteq Q_{D}$
		for every $\mathcal{F}_{i}\in \mathcal{F}$. 
	\end{itemize}
\end{definition}

A run of an LDGBA, subject to an input word $\boldsymbol{w}=w_{0}w_{1}\ldots$, can be expressed as $\boldsymbol{q}=q_{0}q_{1}\ldots$, according to the transition function $\delta(q_i, w_i)=q_{i+1}$. Let $\inf\left(\boldsymbol{q}\right)$ represent the infinite portion of $\boldsymbol{q}$. Theoretically, $\boldsymbol{q}$ satisfies the LDGBA acceptance condition, i.e., the LDGBA accepts the word $\boldsymbol{w}$, if there exists $\inf\left(\boldsymbol{q}\right)\cap \mathcal{F}_{i}\neq\emptyset$, $\forall i\in\left\{ 1,\ldots f\right\}$. We suggest readers check Owl \cite{Kretinsky2018} for more details about automaton generation. This study aims to solve the POMDP problems with LTL constraints as defined below.

\begin{problem}\label{prob1}
   {Given a PL-POMDP $\mathcal{P}$ and a complex task expressed via an LTL formula. The objective is to find a policy $\xi^{*} (\mathbf{o}_t)$, where $\mathbf{o}_t$ denotes a sequence of observations on $\mathcal{P}$, that can complete the task by satisfying the acceptance condition of the LTL-induced LDGBA.}
\end{problem}

\subsection{Product POMDP} \label{sec3c}

\begin{definition}[Product POMDP] \label{def:prodPOMDP} 
Given a PL-POMDP $\mathcal{P}=\left(S,A,T,s_0, R,O, \Omega, \Pi,L\right)$ and an LDGBA $\mathcal{A}=(Q,\Sigma,\delta,$ $q_{0},\mathcal{F})$, the product PL-POMDP (simply named as the product POMDP) is defined by $\mathcal{P^{\times} } =\mathcal{P} \times \mathcal{A}= (S^{\times} , A^{\times}, T^{\times }, s_0^{\times}, R^{\times}, O, \Omega^{\times}, \mathcal{F^{\times}})$, consisting of the following components.
\begin{itemize}
\item A finite set of labeled states, $S^{\times}=S \times Q $ or $s^{\times}=\langle s, q \rangle \in S^{\times}$ where $s\in S$ and $q \in Q$. 
\item A finite set of actions, $A^{\times} = A \cup \left\{ \epsilon \right\}$. 
\item A transition function, $T^{\times } = S^{\times} \times A^{\times} \times S^{\times} \rightarrow [0,1]$, and \begin{equation}
T^{\times } \left( s^{\times}, a^{\times}, s^{\times'} \right) = \left\{ \begin{array}{cc}
T(s, a^{\times}, s') & q'=\delta \left( q, l \right), l \in L(s') \mbox{ and } a^{\times} \in A \\
1 & a^{\times} \in \left\{ \epsilon \right\} \mbox{ and } q' \in \delta(q, \epsilon) \mbox{ and } s'=s,\\
0 & \mbox{otherwise. }
\end{array}\right.
\label{eq:product POMDP transition}
\end{equation}
where ${s^{\times}}' = \langle s', q' \rangle$.
\item An initial state $s_0^{\times} = \langle s_0, q_0 \rangle \in S^{\times}$ where $s_0 \in S$ and $q_0 \in Q$. 
\item A reward function $R^{\times} = S^{\times} \times A^{\times} \times S^{\times} \rightarrow \mathcal{R}$, and
\begin{equation} 
R^{\times} (s^{\times}, a^{\times}, s^{\times'}) = \left\{ \begin{array}{cc} R(s,a^{\times},s')  & a^{\times} \in A, l \in L(s'), q'=\delta \left( q, l \right) \in \mathcal{F}_{i}, \mbox{ and } \mathcal{F}_{i} \in \mathcal{F}\\ 
0 & \mbox{otherwise. }\end{array}\right.
\label{eq:product POMDP reward}
\end{equation}
\item An observation function $\Omega^{\times}=S^{\times} \times A^{\times} \times O \rightarrow [0,1]$ is 
\begin{equation}
\Omega^{\times}(s{^{\times}}', a^{\times},o) =  \Omega(s',a^\times,o)
\label{eq:product POMDP observation}
\end{equation}
If $a^{\times} \in A$. Otherwise, if $a^{\times} \in \{\epsilon\}$, the agent stays at the same state $s$, i.e., $s' = s$, but $q'=\delta(q, \epsilon)$, and no observation is perceived. 
\item A set of accepting sets $\mathcal{F^{\times}} = \left\{ \mathcal{F}_{1}^{\times}, \mathcal{F}_{2}^{\times}, ... , \mathcal{F}_{f}^{\times} \right\}$ where $\mathcal{F}_{i}^{\times} = \left\{ \langle s,q \rangle \vert s \in S; q \in \mathcal{F}_{i} \right\}$ and $i = 1, ... f$. 
\end{itemize}
\end{definition}

A random path on the product POMDP, represented by $(s_0, q_0)(s_1, q_1) \dots$, is an integration of a path  $s_0 s_1 \dots$ on the PL-POMDP and a path $q_0 q_1\dots$ on the LDGBA. Similar to (\ref{eq:expReturn}), the expected return starting from the initial state under a policy $\xi^{\times}$ on the product POMDP can be written as 
\begin{equation}\label{eq:expReturnProductBelief} 
U^{\xi^{\times}} (s_0^{\times}) = \mathbb{E}^{\xi^{\times}} \left[\sum_{t=0}^{\infty} \gamma^t  R(s^{\times}_{t}, a^{\times}_{t}, s^{\times}_{t+1}) \Big \vert s^{\times}_{t=0} =s_0^{\times} \right]
\end{equation}

It shall be noted that the product POMDP $\mathcal{P^{\times}}$ can be viewed as a PL-POMDP $\mathcal{P}$ with the augmented state space, which includes automaton state space. Therefore, the product POMDP accounts for the temporal logic specifications represented by LDGBA $\mathcal{A}$. Any feasible path on $\mathcal{P^{\times}}$ shares the intersections between an accessible path over the original PL-POMDP $\mathcal{P}$ and a word accepted by the LTL-induced automaton $\mathcal{A}$. 
For example, a path $\sigma^{\xi^{\times}}$ = $(s_0, q_0)(s_1, q_1) \dots$ can be generated by the derived policy $\xi^{\times}$ on the product POMDP $\mathcal{P^{\times}}$. If there exists $\inf(\sigma^{\xi^{\times}}) \cap \mathcal{F}_{i}^{\times} \neq \emptyset, \forall i = 1, ... f$, where $\mathcal{F^{\times}}$ captures the acceptance conditions of LDGBA $\mathcal{A}$, this path is accepted. In other words, the run $q_0 q_1 \ldots$ satisfies the LDGBA acceptance condition, or the LDGBA accepts the corresponding word. 

Therefore, according to many previous studies on solving MDP problems with LTL specifications via the product MDP \cite{Cai2021a, Cai2021b, Cai2021c, Bozkurt2020}, an optimal policy ${\xi^{\times}}^{*}(\mathbf{o}_t, \mathbf{q}_t)$ on the product POMDP $\mathcal{P^{\times}}$ is equivalent to the optimal policy $\xi^{*}(\mathbf{o}_t)$ on the PL-POMDP $\mathcal{P}$ while satisfying LTL specifications. $\mathbf{o}_t$ is the observation history while $\mathbf{q}_t$ is the corresponding history of transitioned automaton state. It shall be noted that it is assumed that the agent receives labels, i.e., input symbols in automata, as part of the feedback. Therefore, if the agent is fully aware of the task, i.e., LTL-induced automaton, the history of transitioned automaton states can be derived. Then, we can reformulate Problem \ref{prob1} as follows. 

\begin{problem}\label{prob2}
    A product POMDP $\mathcal{P^{\times} }=\mathcal{P} \times \mathcal{A}$, defined in Section~\ref{sec3c}, is constructed from a PL-POMDP $\mathcal{P}$ describing the partially observable environment and an LDGBA $\mathcal{A}$ expressing LTL specifications $\phi$ for a complex task. The objective is to find a policy ${\xi^{\times}}^{*}(\mathbf{o}_t, \mathbf{q}_t)$, where $\mathbf{o}_t$ and $\mathbf{q}_t$ denote the sequences of observations and transitioned-automaton states, respectively, over $\mathcal{P^{\times} }$ so that the expected return~(\ref{eq:expReturnProductBelief}) is maximized.
\end{problem}

On the other hand, if the agent is unaware of the task, the agent cannot derive automaton state transitions based on the label feedback. Consequently, the policy on the product POMDP is a function of the observation history and the perceived label history as ${\xi^{\times}}^{*}(\mathbf{o}_t, \mathbf{l}_t)$. Then, the above problem formulation can be corresponding revised.

\section{Methodology}\label{sec4}

In this study, we propose model-free RL approaches (i.e., DQNs) on product POMDPs to synthesize optimal motion planning for the agent in a partially observable environment subject to LTL specifications. As discussed and demonstrated in Section~\ref{sec2}, RNNs have the advantage of being included in Q networks for solving POMDP problems because they can capture relative temporal dependencies in the observation history. In the proposed methods, we extend the RNN-enhanced Q networks to process the perceived observations and the recognition of complex tasks. 

We consider two scenarios, depending on whether the agent acknowledges the assigned task. If the agent is explicitly assigned the task, it has full knowledge of the LTL-induced automaton, including the transition function. Consequently, once it reaches a product POMDP state, i.e., an augmented state, it can derive the associated automaton state, although the POMDP state is not fully observable. It shall be noted the state labels are distinguished from the observations. We assume that the agent can acquire state labels correctly via feedback from the environment. Therefore, in this case, in addition to the observation history, our method takes the identified automaton state as another input feature to Q networks. On the other hand, if the agent is not explicitly assigned the task and has no knowledge of the automaton's transition function, (POMDP) state labels are taken as an additional input feature to Q networks.

\subsection{Automaton state sequence as additional input}\label{sec4a}

Suppose the agent is fully aware of the task. In that case, the automaton is fully knowledgeable, and the automaton states (i.e., $q$ states) can be induced from the acquired labels according to the automaton transition function. Consequently, in addition to the observation history, automaton states can be directly used as input to the Q networks. Indeed, we utilize a sequence of $q$ states with a length of $k$, $\mathbf{q}_t= {\bar{q}}_1 \dots {\bar{q}}_k$, where $\bar{q}$ denotes the induced $q$ state, representing the automation evolution by times step $t$ corresponding to the agent's transitions on the PL-POMDP. It shall be noted that the sequences of observations $\mathbf{o}_t= o_{t-j} \dots o_t$ have a length of $j+1$ and are generated based on the observation history from time step $t-j$ to the current time $t$. However, since the automaton space is usually smaller than the POMDP space, and the automaton state transition doesn't occur at each time step, it is not practical to generate the history of the automaton transition by recording the automaton state at each time step. Instead, we use First in, first out (FIFO) to track the automaton transitions, and each transition is recorded only once. Consequently, $\mathbf{q}_t$ may maintain the same for several consecutive steps until the next automaton state is reached. In addition, $\mathbf{o}_t$ and $\mathbf{q}_t$ usually don't have the same length.

After each transition $a^{\times}_t \in A$, the observed $o_{t+1}$ and induced automaton state $q_{t+1}$ are used to generate the new sequences of observations $\mathbf{o}_{t+1}$ and $q$ states $\mathbf{q}_{t+1}$. Together with the previous sequences, a new experience can be written as $(\mathbf{o}_t, \mathbf{q}_t, a^{\times}_t, r^{\times}_t, \mathbf{o}_{t+1}, \mathbf{q}_{t+1})$, which is recorded as one data sample in the replay memory. If there is no new automation transition, $\mathbf{q}_{t+1}=\mathbf{q}_{t}$. It shall be noted that the $\epsilon$-transition is one of the available actions in the product POMDP as defined in Section~\ref{sec3c} if it exists in the LTL-induced LDGBA. In our approaches, no observation is perceived after an $\epsilon$-transition because the agent remains at the same POMDP state. However, the automaton state is changed and recorded in the corresponding automaton state sequence for the next time step. 

\begin{figure}[ht]%
\centering
\includegraphics[width=130mm]{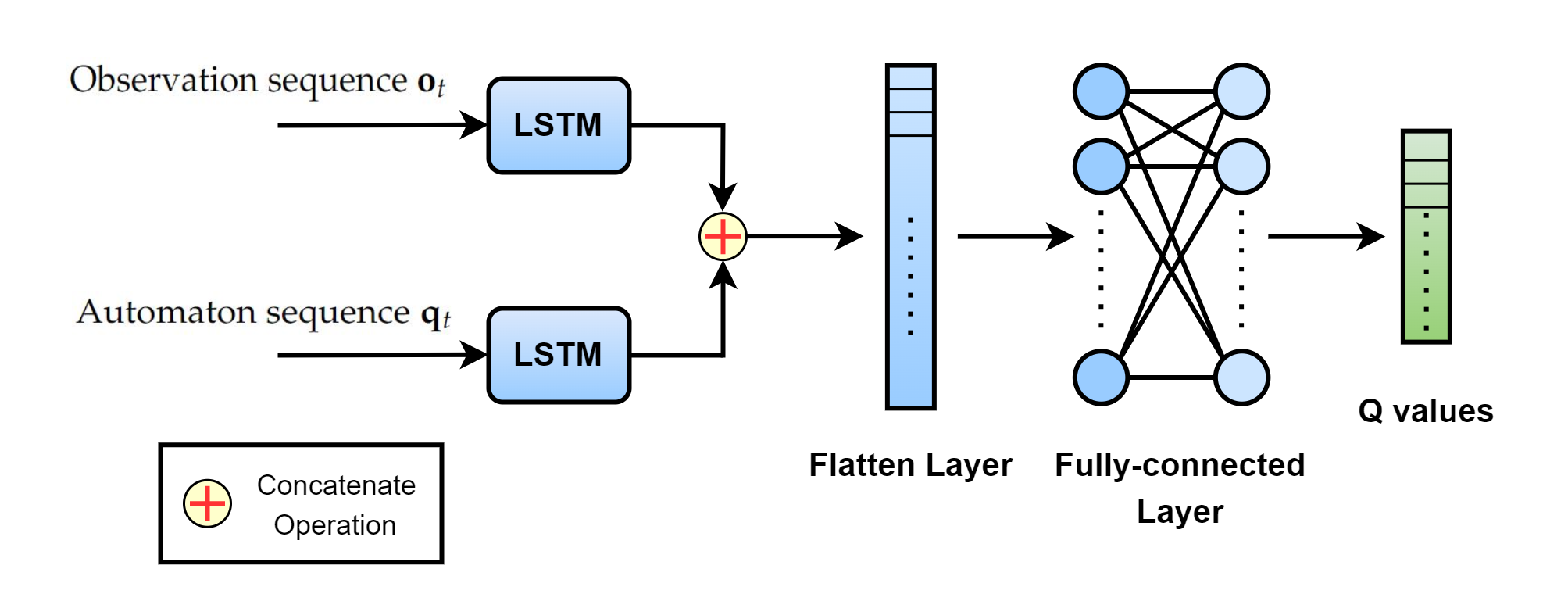}
\caption{The architecture of Q networks taking $\mathbf{o}_t$ and $\mathbf{q}_t$ as input.}\centering\label{fig:DQN with RNN architecture 2}
\end{figure}

Figure~\ref{fig:DQN with RNN architecture 2} illustrates the architecture of the Q networks in our model-free RL approaches. The sequences of observations $\mathbf{o}_t$ and automaton states $\mathbf{q}_t$ at time step $t$ are pre-processed by one-hot-encoding before entering LSTMs, respectively. Since these two sequences don't have the same length, two separate LSTMs are adopted to extract the hidden states, which are then concatenated and flattened for the fully-connected layers to estimate Q values.

Similar to a general DQN, our DQN on the product POMDP also has two identical Q networks: the evaluation network $Q^\times_E(\mathbf{o}_t, \mathbf{q}_t, a^\times_t; \theta^\times_E)$ and the target network $Q^\times_T(\mathbf{o}_t, \mathbf{q}_t, a^\times_t; \theta^\times_T)$ where $a^\times \in A$. The target network is utilized for the next action selection and Q value prediction. On the other hand, the evaluation network is trained every $M$ steps by a batch of data samples randomly selected from the replay memory. After every $K$ steps, the target network is updated by copying the weight coefficients of the evaluation networks. Using two neural networks can prevent the bootstrapping of the DQN with a single neural network. 

The output of each data sample in the selected batch is a Q value, which can be updated as the equation below.
\begin{equation}\small
Q^\times_{\text{new}}(\mathbf{o}_t,\mathbf{q}_t, a^{\times}_t) = Q^\times_E(\mathbf{o}_t,\mathbf{q}_t,a^{\times}_t; \theta^\times_E)+\alpha \left[ r^{\times}_t+\gamma \max_{a^{\times}_{t+1}} Q^\times_T(\mathbf{o}_{t+1},\mathbf{q}_{t+1},a^{\times}_{t+1}; \theta^\times_T)-Q^\times_E(\mathbf{o}_t,\mathbf{q}_t,a^{\times}_t; \theta^\times_E) \right]\label{eq:DRQN Q value on product}
\end{equation}
where $r^{\times}_t=R^{\times}(s^{\times}_t, a^{\times}_t, s^{\times'}_t)$, $\alpha$ is the learning rate, and $\gamma$ is the discount factor. Therefore, each data sample has the input features $\mathbf{o}_t$ and  $\mathbf{q}_t$ and the output target $Q^\times_{\text{new}}$ to train and update the evaluation Q network, $Q_E$. In addition, the following loss function is used to update the weight coefficients of the evaluation Q-network.
\begin{equation}\label{eq:DRQN_loss_function on product}\small
\mathcal{L}(\theta^\times_E) = \mathbb{E}_{(\mathbf{o}_t,\mathbf{q}_t,a^{\times}_t,r^{\times}_t,\mathbf{o}_{t+1},\mathbf{q}_{t+1})\sim U(D)} \left[ \left(r^{\times}_t+\gamma \max_{a^{\times}_{t+1}} Q^\times_T(\mathbf{o}_{t+1},\mathbf{q}_{t+1}, a^{\times}_{t+1}; \theta^\times_T) - Q^\times_E(\mathbf{o}_t,\mathbf{q}_t, a^{\times}_t; \theta^\times_E)\right)^2 \right]
\end{equation}

Algorithm~\ref{algo1} demonstrates training the evaluation Q network during the agent interactions with the environment. Once converged, the Q network can predict the state-action value (i.e., Q value) function to derive the optimal policy for the studied POMDP problem with temporal logic specifications.

\begin{algorithm}
\caption{Deep Recurrent Q-Network for Product POMDP Problems.}\label{algo1}
\begin{algorithmic}[1]
\State Initialize LTL formula $\phi$, POMDP $\mathcal{P}$.
\State Convert $\phi$ to an LDGBA $\mathcal{A}$.
\State Construct the product POMDP $\mathcal{P^{\times} }=\mathcal{P} \times \mathcal{A}$.
\State Initialize the evaluation network $Q^\times_E$, the target network $Q^\times_T$, the replay memory $D$, the length of observation sequence $j+1$, the empty $q$ state sequence $\mathbf{q}_{0}$ with length of $k$, the learning rate $\alpha$, the discount factor $\gamma$, the total number of episodes $E$, the total number of steps $N$, the batch size $M$, and the $Q^\times_T$ update steps $K$. 
\While{The current episode $e$ in $E$} 
    \State Randomly select a start state $s^{\times}_0$.
    \While{The current step $i$ in $N$}
        \State Select a random action $a^{\times}_i$ if $i < j+1$; otherwise, select an action via the $\epsilon$-greedy technique. 
        \State Obtain the observation $o_{t+1}$ and induce the automaton state $q_{t+1}$.
        \State Generate $\mathbf{o}_{t+1}$ and $\mathbf{q}_{t+1}$.
        \State Collect the rewards $r^{\times}_i$. 
        \State Store the experience $\langle \mathbf{o}_i,\mathbf{q}_i,a^{\times}_i,r^{\times}_i,\mathbf{o}_{i+1},\mathbf{q}_{i+1} \rangle$ in  $D$. 
        \If{$i > 0$ and $i\%M$=0}
            \State Randomly select $M$ data samples as U($D$) from the replay memory. 
            \State Compute $Q^\times_{new}$ for each data sample.
            \State Train $Q^\times_E$ by the batch of samples.
        \EndIf
        \If{$i > 0$ and $i\%K$=0}
            \State Pass the weights of $Q^\times_E$ to $Q^\times_T$. 
        \EndIf
    \EndWhile
\EndWhile
\State Training end and save the evaluation network $Q^\times_E$
\end{algorithmic}
\end{algorithm}

\subsection{Label sequence as the additional input}\label{sec4b}
If the agent is unaware of the task, i.e., the LTL-induced automaton is not knowledgeable, the state labels need to be the additional input for Q networks. We modified the Q network architecture in Figure \ref{fig:DQN with RNN architecture 2} by replacing the automaton state sequence with a label sequence (i.e., a sequence of input symbols) in addition to the observation sequence. As mentioned above, state labels differ from observations and can be precisely received by the agent as feedback. Consequently, the sequence of the collected experience becomes $(\mathbf{o}_t, \mathbf{l}_t, a^{\times}_t, r^{\times}_t, \mathbf{o}_{t+1}, \mathbf{l}_{t+1})$, where $\mathbf{l}_t$ is the sequence of labels with a length of $k$ received by the time step $t$. Similar to the scenario of utilizing $q$ states, the labels corresponding to the POMDP states can be sparse. Hence,  we utilize the same FIFO method (as described in Section \ref{sec4a}) to generate the label sequence that only stores a label once it is received. On the other hand, Algorithm~\ref{algo1} can be corresponding revised. 

\section{Simulations and Results}\label{sec5}

We evaluate our approaches on three simulations with discrete POMDP domains. We first perform simulations over a partially observable grid world with two different tasks, considering $\epsilon$-transition in the automaton and static/dynamic events in the PL-POMDP. Then, we test the approaches in an office scenario where we utilize two different observation settings. Finally, we also conduct a preliminary application of the proposed approach to a multiagent RL case. The simulations are programmed via Python 3.9 and Rabinizer 4. They are completed on a desktop with a 3.20 GHz eight-core CPU and 32 GB RAM. Part of the source codes and supplementary materials are provided \footnote{\url{https://github.com/JunchaoLi001/Model-free_DRL_LSTM_on_POMDP_with_LDGBA}}.

\subsection{Grid world Simulations}\label{sec5a}

\begin{figure}
\centering
\includegraphics[width=50mm]{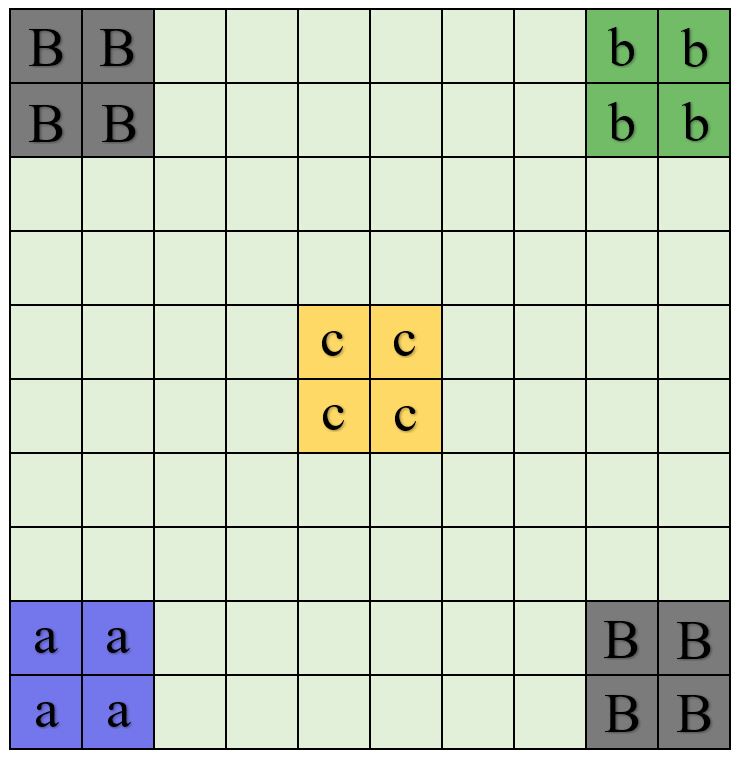}
\caption{A 10 x 10 grid world with trapping states}\centering\label{fig:grid-world_2}
\end{figure}

We use a 10 × 10 grid world workspace, shown in Figure.\ref{fig:grid-world_2}. Several states are labeled with `a' in blue and `b' in green, indicating two different events. The trapping states, labeled with `c', indicate that the agent can never leave once entering them. Other parameters, like the transition probability and observation probability, are the same as defined in the simple go-to-goal case in Section~\ref{sec2c}. Two tasks are studied. Task 1 demonstrates that our approaches can handle LDGBA with $\epsilon$-transitions. Task 2 considers two scenarios: static events and dynamic events, respectively. When assuming dynamic events, each event has a 90\% probability of occurring at its labeled states and a 10\% probability at the other labeled states. Both tasks are simulated for 15,000 episodes with 600 steps per episode, using the observation sequence with a length of $j+1=5$, the automaton state or label sequence with a length of $k=3$, batch size $M=32$, and the number of steps to copy the evaluation Q network $K=50$. 

\subsubsection{Task 1}\label{sec5aa}

\begin{figure}[ht]%
\centering
\includegraphics[width=80mm]{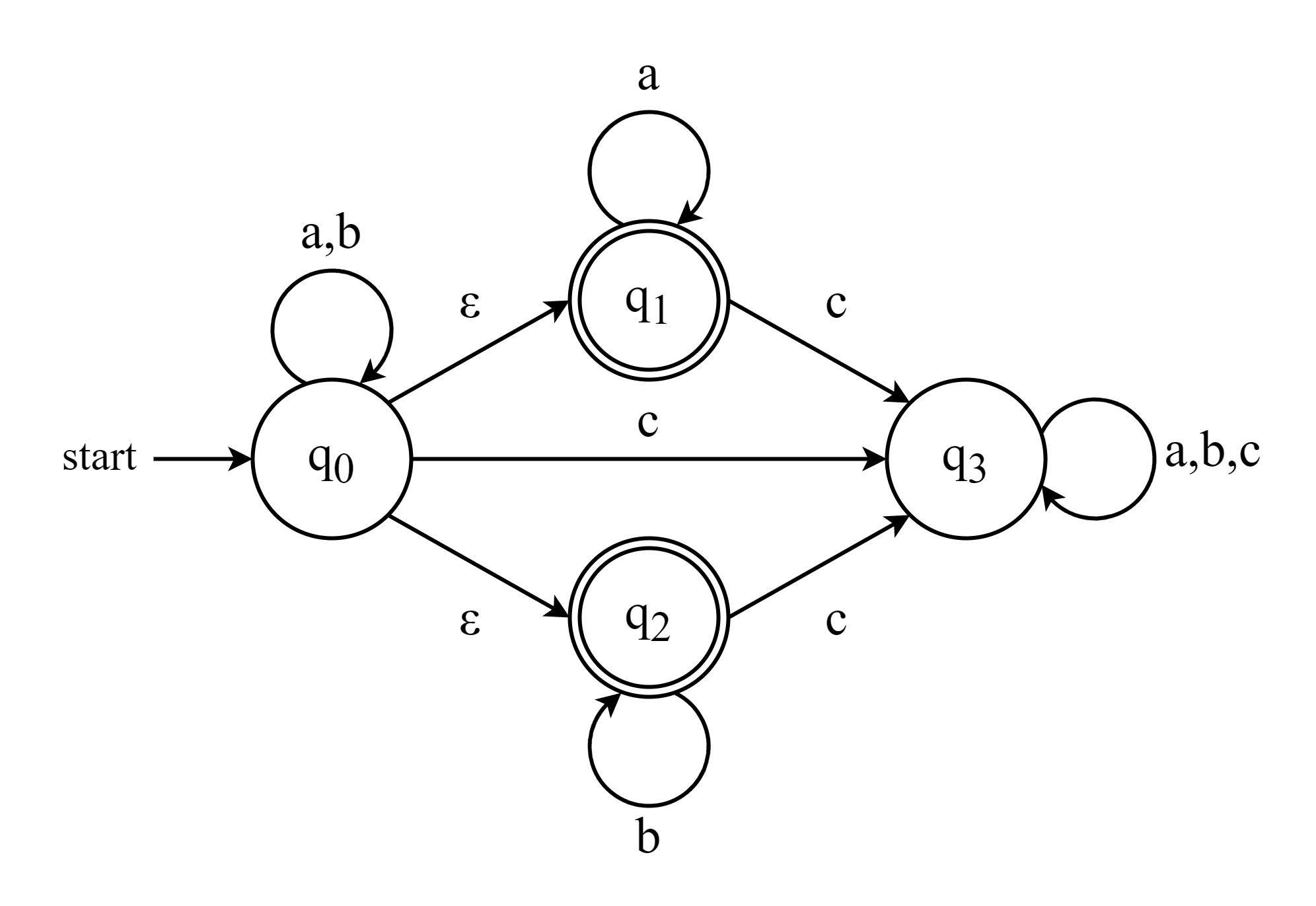}
\caption{The LDGBA of $\phi_1$.}\centering\label{fig:automaton task1}
\end{figure}

The first task tested in the grid world requires the agent to visit states labeled `a' or `b' infinitely many times. The LTL formula is expressed below, and the induced LDGBA is shown in Figure~\ref{fig:automaton task1}.
\begin{equation}
 \phi_1 = (\square \diamondsuit a \mid \square \diamondsuit b) \wedge \square \neg  c 
 \label{eq:gridworld_task1}
\end{equation} 

It can be seen that the LDGBA contains $\epsilon$-transitions that are included in the set of actions, $a^{\times}$, in the generated product POMDP. Figure~\ref{fig:automaton task1} also shows that the $\epsilon$-transitions can transition the automaton states from the initial state $q_0$ to either $q_1$ or $q_2$, resulting in the agent to keep visiting only `a' or `b', respectively. Visiting `c' will lead to a state ($q_3$) without an outlet as a "trapping" state. 

According to the definition of LDGBA in Section~\ref{sec3b},  $\epsilon$-transitions don't take any input symbols. They are only valid to enter the deterministic set of automaton states where the transitions are restricted. Therefore, after an $\epsilon$-transition, the agent will be at the augmented states associated with either $q_1$ or $q_2$ to complete the task. Without the loss of generality, we give each episode a random probability for selecting the $\epsilon$-transition, which only occurs once. Otherwise, the other actions, i.e., $a^\times \in A$, will be chosen based on the $\epsilon$-greedy method. As mentioned in Section~\ref{sec3c}, the agent doesn't perceive observations right after taking $\epsilon$-transitions, and Q networks only predict Q values of the actions other than $\epsilon$-transitions.

\begin{figure}[ht]%
\centering
\includegraphics[width=80mm]{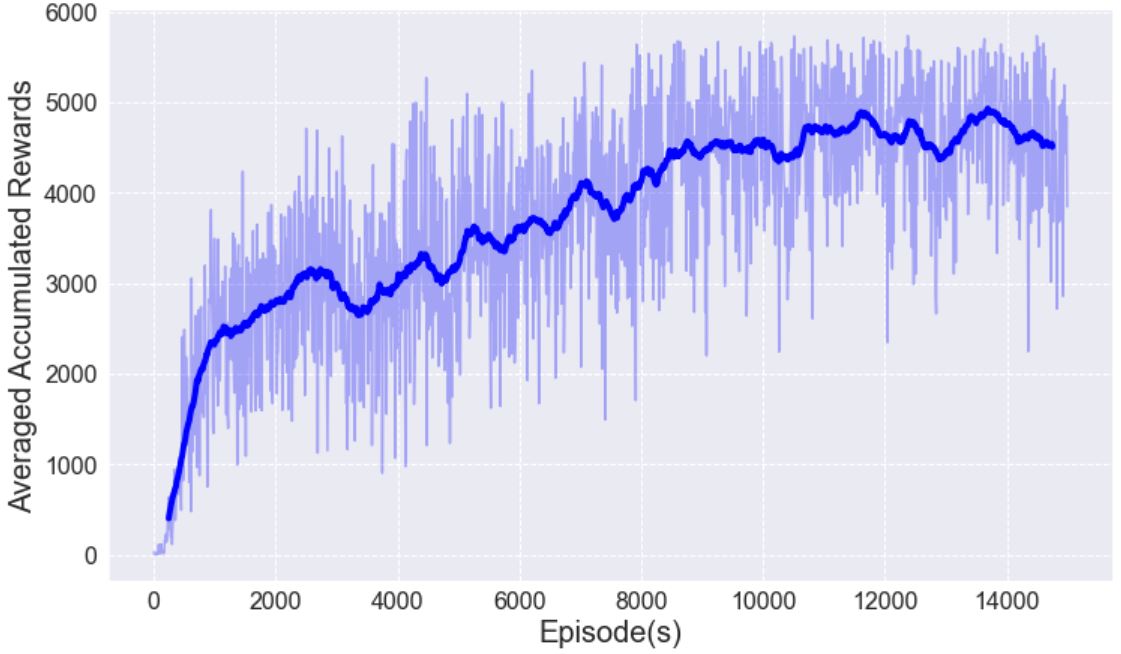}
\caption{The averaged accumulated rewards of task $\phi_1$.}\centering\label{fig:plot grid task epsilon}
\end{figure}

\begin{figure}
  \centering
  \includegraphics[width=100mm]{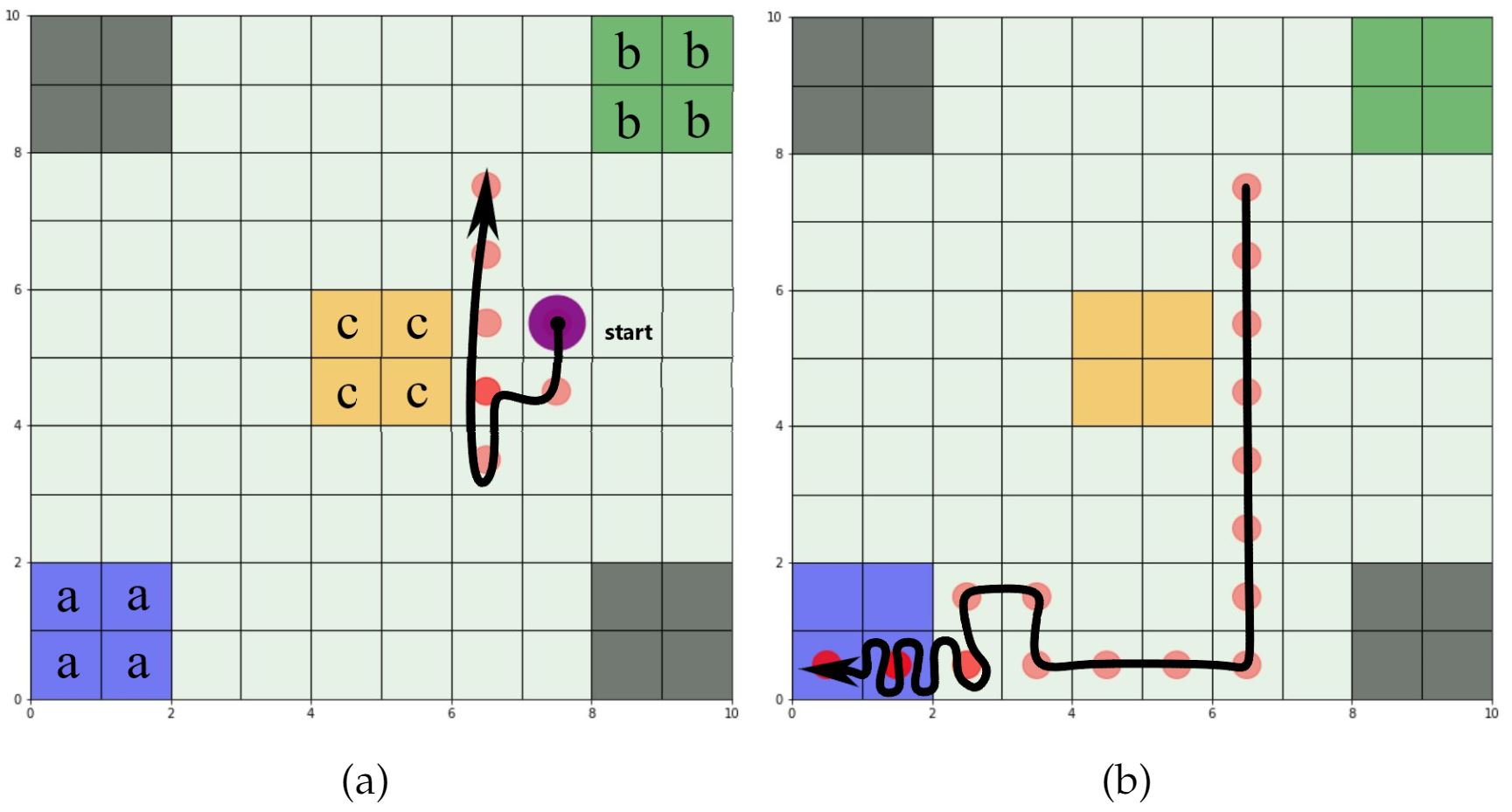}
  \centering\caption{A single round path of task $\phi_1$: (\textbf{a}) The path on $q_0$. (\textbf{b}) The path on $q_1$.}
  \label{fig:DRQN_task_epsilon_q_state_no_uncertainty}
\end{figure}

In Task 1, we only consider static events, i.e., $P_L(s_a, `a')=P_L(s_b, `b')=1$. Also, it is assumed that the agent is aware of the task. Therefore, the observation and $q$ state histories are input to predict Q values via Q networks for action selection. Figure~ \ref{fig:plot grid task epsilon} shows the evolution of accumulated reward, averaged per 10 episodes with SMA 50 episodes. Since the agent may accidentally move into the `trapping' state, averaged rewards better presents the trend of the rewards' convergence. The reward is set as 10 whenever the agent visits accepting states. After obtaining the optimal policy, we generate one path for the agent to accomplish the task as shown in Figure~ \ref{fig:DRQN_task_epsilon_q_state_no_uncertainty}. 

At first, the agent randomly selects actions, as shown in Figure~ \ref{fig:DRQN_task_epsilon_q_state_no_uncertainty}(a), before generating the first observation and $q$ state sequences. Then, Q values can be predicted, and the greedy action selection is applied unless an $\epsilon$-transition is taken. After the $\epsilon$-transition, the automaton state is transitioned from $q_0$ to $q_1$, and the agent desires to visit the state labeled `a' infinitely many times. This can be seen in Figure~\ref{fig:DRQN_task_epsilon_q_state_no_uncertainty}(b), where the agent moves down and then left to keep visiting state `a'. Due to the action uncertainty, the agent occasionally visits some states multiple times, indicated as bright red dots in Figure~\ref{fig:DRQN_task_epsilon_q_state_no_uncertainty}.

\subsubsection{Task 2}\label{sec5ab}

The second task requires the agent to visit states labeled `a' then `b' in order infinitely many times, subject to dynamic events due to labeling uncertainty. The LTL formula is expressed as 
\begin{equation}
\phi_2 = \square \diamondsuit (a \wedge \diamondsuit b) \wedge \square \neg  c 
\label{eq:gridworld_task2}
\end{equation}

\begin{figure}[ht]%
\centering
\includegraphics[width=60mm]{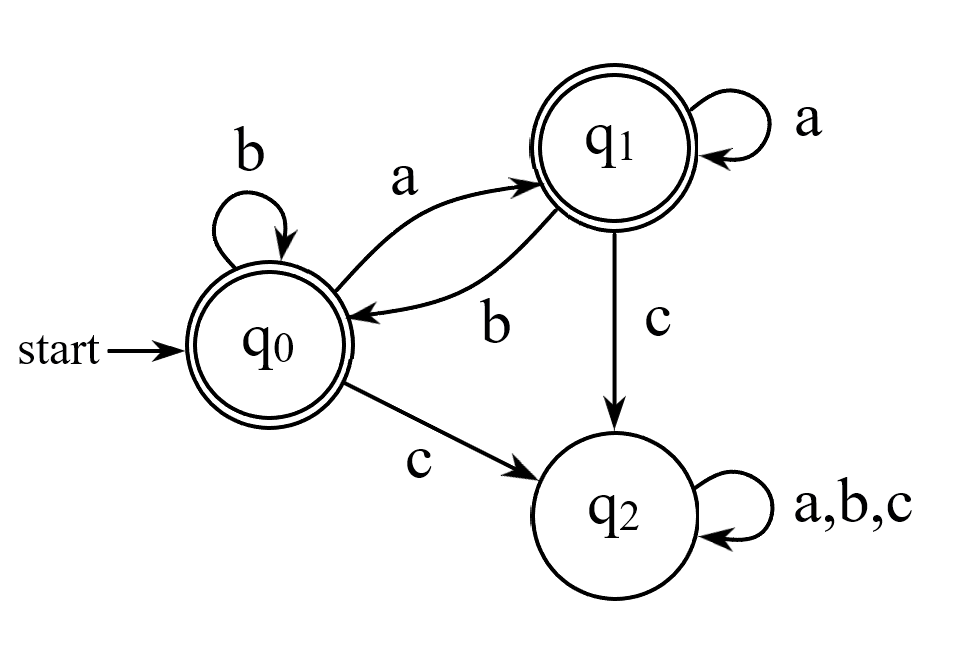}
\caption{The LDGBA of $\phi_2$.}\centering\label{fig:automaton task2}
\end{figure}

In the LTL-induced LDGBA, directly utilizing the accepting sets may fail to find the deterministic policy as discussed in \cite{Hasanbeig2019, Oura2020}. Inspired by their works, we modify the automaton structure and reward function for easing the training process. Specifically, we augment the accepting states to separate transitions with the input symbols `a' or `b', respectively. Only single-labeled transitions are kept for demonstration, as shown in Figure \ref{fig:automaton task2} for Task 2. We then redesign the reward function (\ref{eq:product POMDP reward}), shown below, by adding a constraint to the reward function so that the agent can visit the accepting sets repeatedly. 
\begin{equation} 
R^{\times} (s^{\times}, a^{\times}, s^{\times'}) =
\left\{ \begin{array}{cc} R(s,a^{\times},s')  & a^{\times} \in A, l \in L(s), q'=\delta \left( q, l \right) \in \mathcal{F}_{i}, \mathcal{F}_{i} \in \mathcal{F}, \mbox{ and } q' \neq q \\ 
0 & \mbox{otherwise. }\end{array}\right.
\label{eq:product POMDP reward redesign}
\end{equation}
where $q \neq q'$ prevents the repeated transitions at the same automaton accepting state by removing the rewards on the associated labeled POMDP states. After applying this constraint to the reward function, the derived optimal policy satisfies the desired surveillance task specification. An alternative approach can be implementing the frontier-tracking function introduced by M. Cai \textit{et al.} \cite{Cai2021b}.

For Task 2, we investigate the agent’s learning in the environment with dynamic or static events combined with two scenarios that depend on the agent’s awareness of the task. At first, to provide a detailed demonstration, we consider dynamic events and assume that the agent fully understands the task. The labeling uncertainty is introduced so that an event has a 90\% probability of occurrence at its labeled states and the other labeled states otherwise. For example, at the states labeled ‘a’, event `a' has a 90\% probability of occurrence while event ‘b’ has a 10\% probability. In addition, a $q$ state sequence is utilized as the input of Q networks in addition to the observation sequence. 

\begin{figure}[ht]%
\centering
\includegraphics[width=130mm]{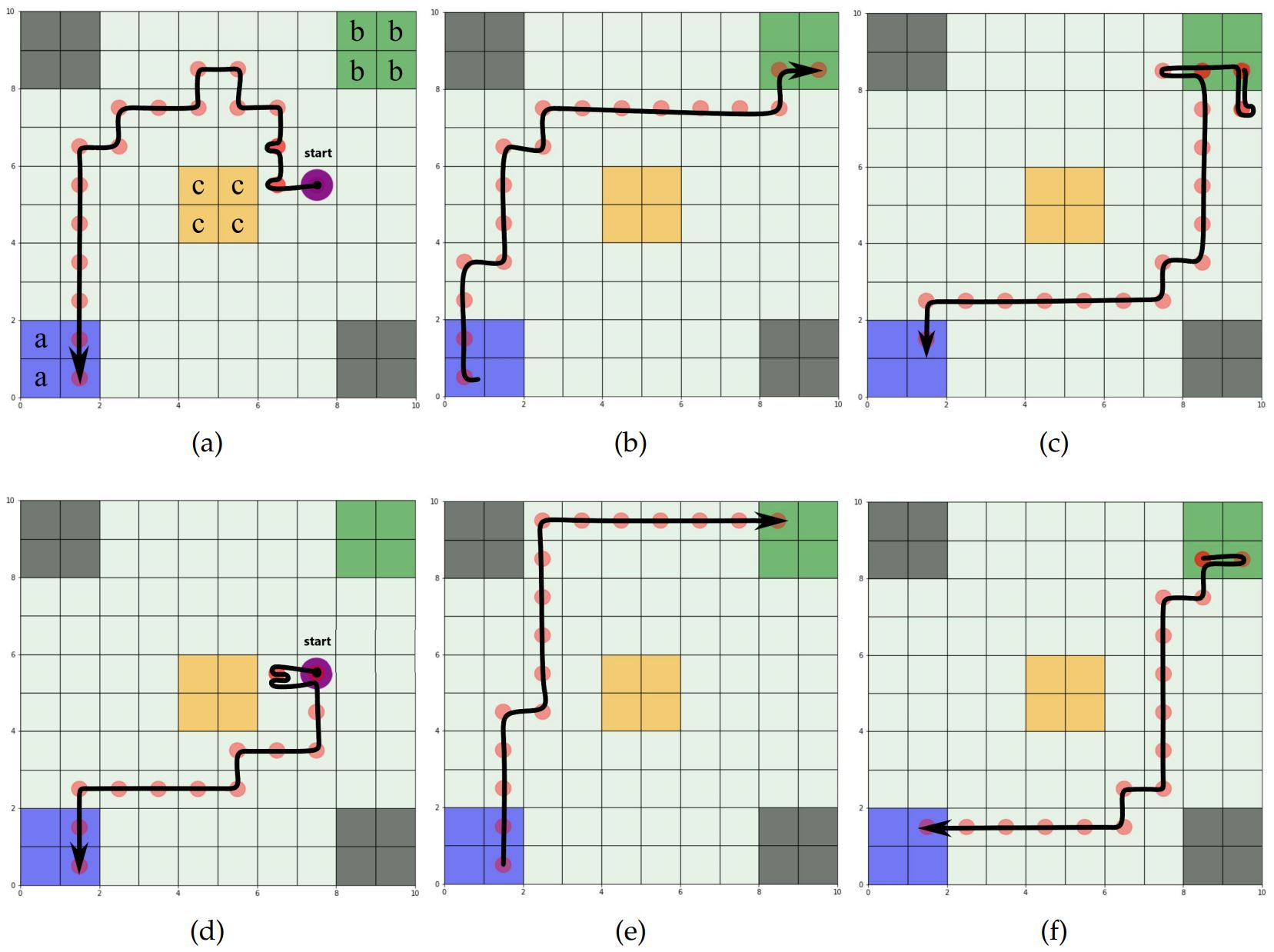}
\caption{A generated path for Task in $\phi_2$ of dynamic event with labeling uncertainty: \\ \textbf{(1)} If it is fully aware of the task ($q$ state sequence): (\textbf{a}) The path on $q_0$. (\textbf{b}) The path on $q_1$. (\textbf{c}) The path back on $q_0$.\\ \textbf{(2)} If it is not aware of the task (label sequence): (\textbf{d}) The path on $q_0$. (\textbf{e}) The path on $q_1$. (\textbf{f}) The path back on $q_0$.}\centering
\label{fig:DRQN_task2_q_state_with_uncertainty}
\end{figure}

Figure~\ref{fig:DRQN_task2_q_state_with_uncertainty}(1) demonstrates a single round path generated from the learned policy for the agent to accomplish Task 2 when the agent is fully aware of the task. In Figure~\ref{fig:DRQN_task2_q_state_with_uncertainty}(a), the agent starts from the initial state (purple dot) and performs five random movements on $q_0$ until it generates the first observation sequence (along with the q state sequence) for decision-making. It then navigates around the trapping states and heads to the bottom left for the states labeled `a' shown as the blue square. However, event `a' doesn't occur in the state when the agent first visits the blue area due to the labeling uncertainty. Therefore, the agent moves downward to the next state, where event `a' occurs at the next time step. The bend of the black route at the top of the path is caused by motion uncertainty.

After visiting the state labeled `a', the automaton transition happens from $q_0$ to $q_1$. In Figure~\ref{fig:DRQN_task2_q_state_with_uncertainty}(b), the agent moves around the area of `a' states then bypasses the trapping states in yellow to reach the states labeled `b' in the top right corner. Again, event `b' doesn't occur on the agent's first visit. Then, the agent moves one more step to the right, and event `b' occurs. Consequently, the agent completes a single round to visit `a' and then `b'. At the same time, the agent is back to the automaton state $q_0$. Figure~\ref{fig:DRQN_task2_q_state_with_uncertainty}(c) shows the agent tries to move back to visit states labeled `a' for the second round, but it keeps visiting `b' states a few times before heading to `a' states.

We also conduct the simulation in an environment with dynamic events when the agent is unaware of the task. The observation and label sequences are input to the Q networks in this case. Figure~\ref{fig:DRQN_task2_q_state_with_uncertainty} (2) illustrates the path generated from the learned policy. We observe a similar phenomenon in which the agent makes a few attempts to visit `a’ states before leaving for `b’ states. However, after we generate more paths, we find that the agent occasionally visits the states labeled `b’ first because event `a’ has 10\% probability of occurring on those states. This phenomenon is uniquely observed when the agent is unaware of the task. 

\begin{figure}[h]%
\centering
\includegraphics[width=130mm]{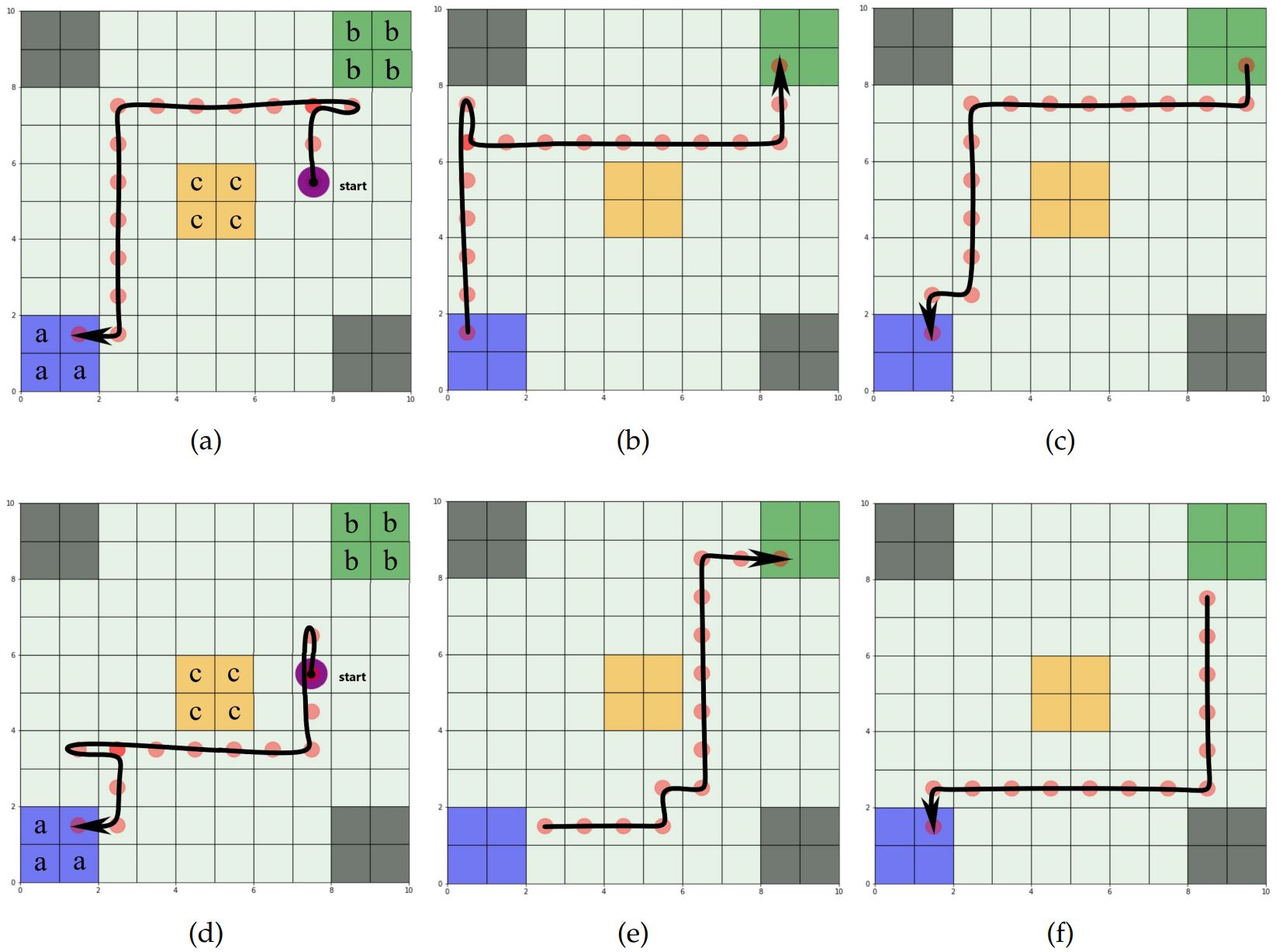}
\caption{A generated path for Task in $\phi_2$ of static event:
\\ \textbf{(1)} If it is fully aware of the task ($q$ state sequence): (\textbf{a}) The path on $q_0$. (\textbf{b}) The path on $q_1$. (\textbf{c}) The path back on $q_0$.
\\ \textbf{(2)} If it is not aware of the task (label sequence): (\textbf{d}) The path on $q_0$. (\textbf{e}) The path on $q_1$. (\textbf{f}) The path back on $q_0$.}\centering
\label{fig:grid_task2_q_state_label_no_uncertainty}
\end{figure}

Next, the events are assumed to be static, i.e., without labeling uncertainty. We consider both scenarios depending on whether the agent is fully aware of the task. After obtaining optimal policies for each scenario, we generate paths to demonstrate the agent accomplishing the task. Figure~\ref{fig:grid_task2_q_state_label_no_uncertainty} \textbf{(1)} shows the agent planning the motion based on the observation history and the $q$ state history. On the other hand, Figure~\ref{fig:grid_task2_q_state_label_no_uncertainty} \textbf{(2)} demonstrates that the agent can accomplish the task following the policy regarding the observation history and the label-receiving history. By following both paths, the agent can complete the task, and the performances between the two approaches are similar. Comparing the dynamic event cases, we don’t observe that the agent tries to visit `a’ or `b’ states multiple times before leaving for the other. 

\begin{figure}[h]%
\centering
\includegraphics[width=80mm]{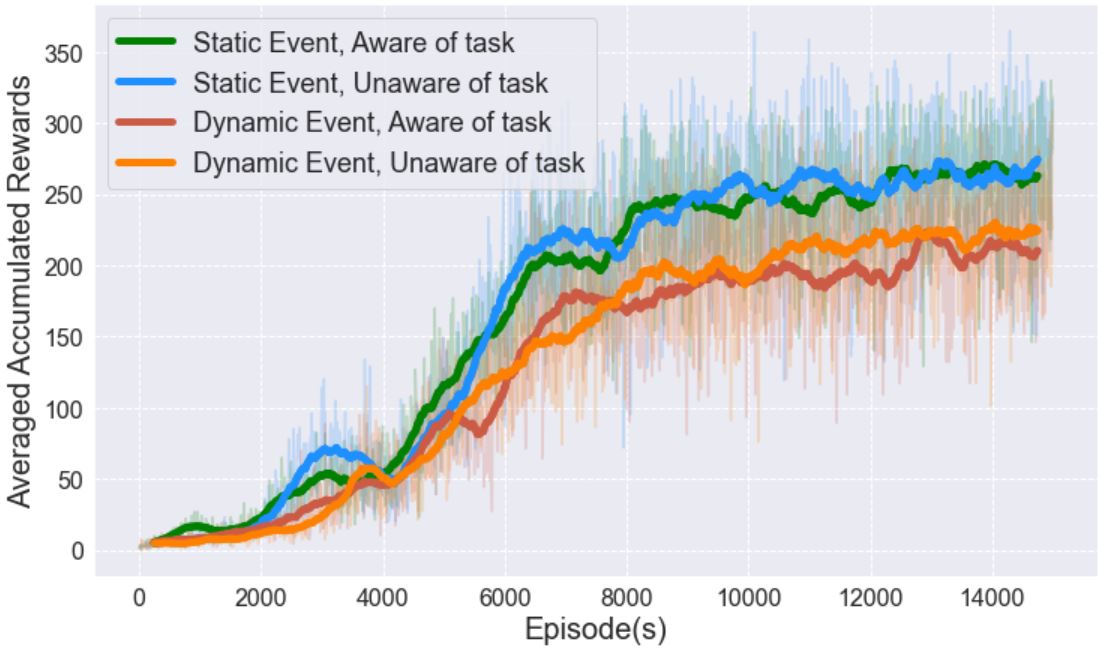}
\caption{Comparison of the reward evolution for task  $\phi_2$.} \centering\label{fig:plot_grid_task2}
\end{figure}

Figure~\ref{fig:plot_grid_task2} compares the evolution of the accumulated rewards of Task 2 for four cases discussed above. It can be observed that static event cases reach higher accumulated rewards than dynamic event cases. It may be due to the labeling uncertainty in dynamic events. Also,  regardless of the agent’s awareness of the task, the converged accumulated rewards are similar.

\subsection{Pybullet TurtleBot Simulations} \label{sec5b}
\begin{figure}[h]%
\centering
\includegraphics[width=65mm]{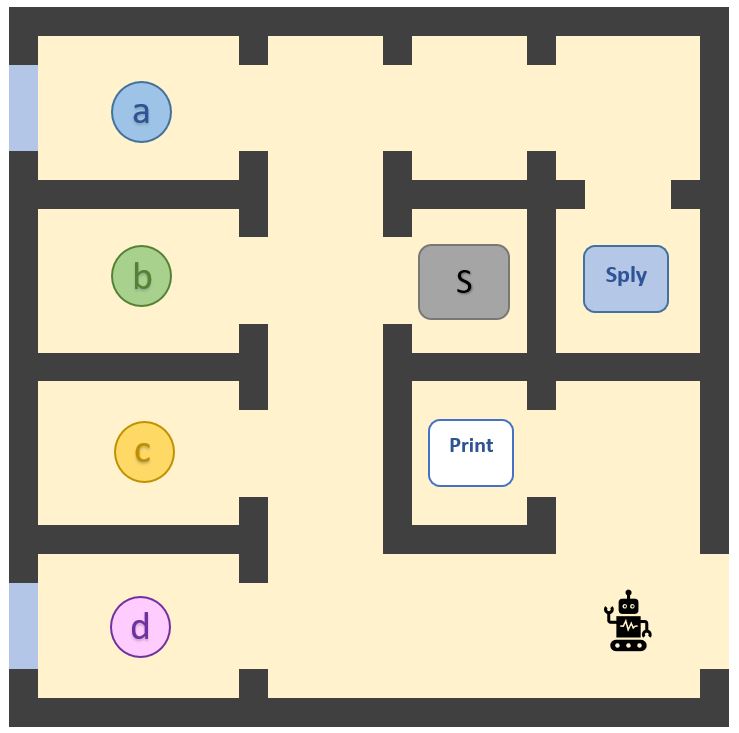}
\caption{The office environment.}\label{fig:office}
\end{figure}

Figure. \ref{fig:office} shows a virtual office environment generated by PyBullet 3.0 \cite{coumans2021}. In the office space, there are four office rooms `a’, `b’, `c’, and `d’, a storage room `S’, a printer room `Print’, and a supply station `Sply'  to recharge the TurtleBot, i.e., the agent. In addition, there are two big windows in Offices `a' and `d' and multiple doors in the office space. We discretize this office space into a four-by-four grid world to generate the POMDP model. Considering the motion uncertainties, we assume that the TurtleBot has a probability of 0.9 to successfully execute its navigation controller by moving along the desired direction. However, it can move to other possible directions, uniformly sharing a probability of 0.1. Moving toward the wall will keep the TurtleBot stay at the same location. Assuming the agent is fully aware of the assigned tasks, We test the proposed model-free RL approach with two different observation settings in this office scenario. All simulations are conducted via 10,000 episodes with 300 steps per episode. The other computation settings are the same as in Section~\ref{sec5a}. We map the office scenario for each simulation to a grid world in which the optimal policy is learned. Then, we apply the derived optimal policy to the virtual TurtleBot in the PyBullet platform to validate the task accomplishment. 

\subsubsection{Observation of the surroundings} \label{sec5ba}
This setting assumes that at the current state, the TurtleBot can collect the surrounding observations in all four directions, following a specific order from `North', `West', `South' to `East'. The observation elements are `wall', `hallway', `door', and `window'. The agent can only observe one element in each direction. It shall be noted that there is only one observation at each state, i.e., $O(s', a,o)=1$. However, the agent may perceive the same observation in two or more states. For example, offices `b' and `c' have the same observation: $o$(`b') = $o$(`c') = \{`wall' `wall' `wall' `door'\}. Consequently, the set of observation $O$ in this POMDP problem consists of 13 distinct observations. 

\begin{figure}[h]%
\centering
\includegraphics[width=120mm]{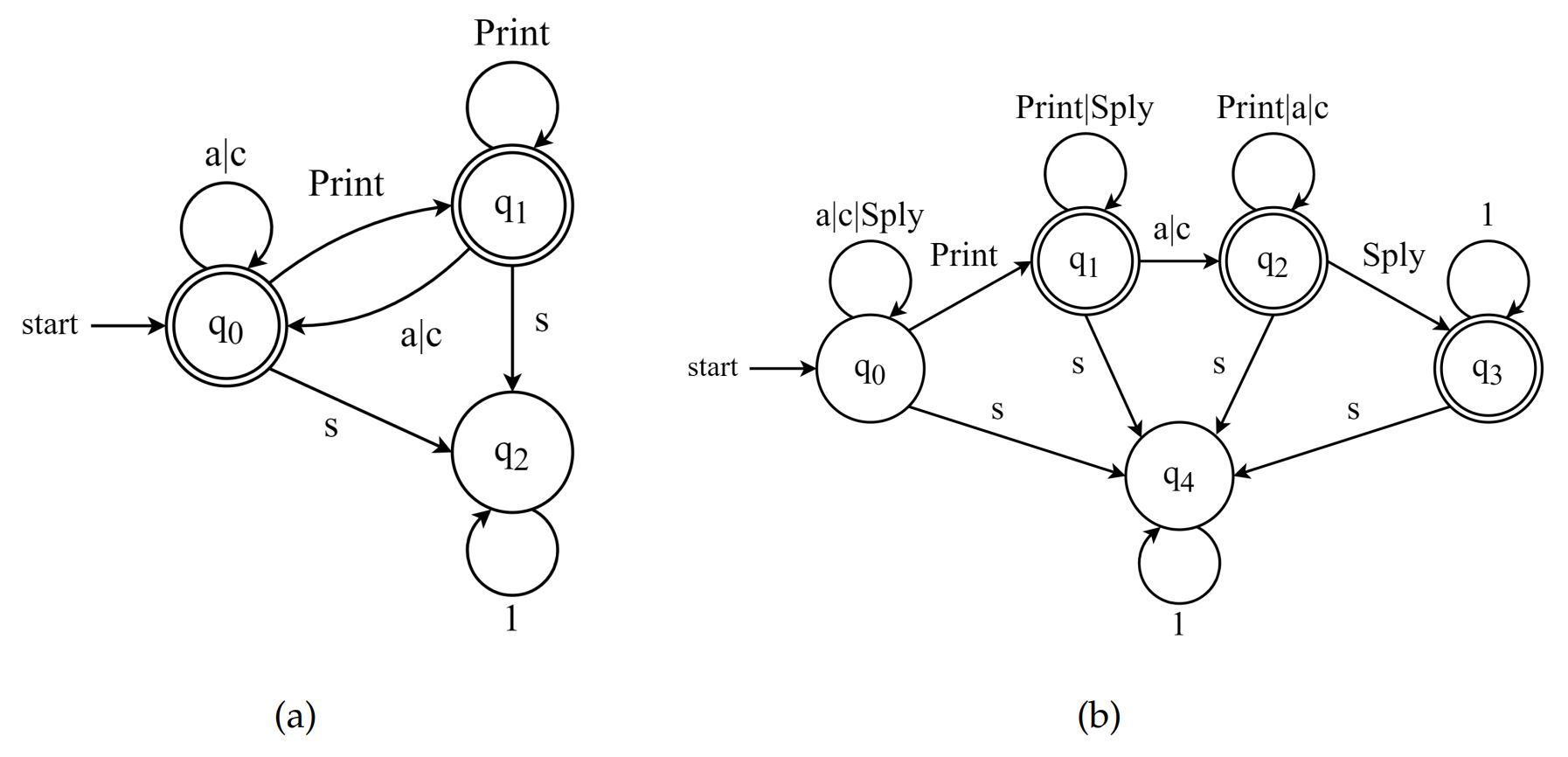}
\caption{The induced LDGBAs: (\textbf{a}) $\varphi_{task1}$. (\textbf{b}) $\varphi_{task2}$.}\centering\label{fig:office_task_automaton}
\end{figure}

\paragraph{Task 1:}
Task 1 requires that the TurtleBot visits the printer room to collect the documents and then carries the documents to Office `a’ or `c’, repeatedly. At the same time, the TurtleBot shall always avoid entering the storage room `S'. Similar to equation~(\ref{eq:gridworld_task2}), this task can be expressed as an LTL formula in equation~(\ref{eq:office_task1}). Figure~\ref{fig:office_task_automaton}(a) shows the induced LDGBA, and we also employ the redesigned reward in equation~(\ref{eq:product POMDP reward redesign}). 

\begin{equation} \label{eq:office_task1}
 \varphi_{task1}= \square \diamondsuit (Print \wedge \diamondsuit (a \mid c)) \wedge \square \neg  S 
\end{equation} 

\begin{figure}[h]%
\centering
\includegraphics[width=100mm]{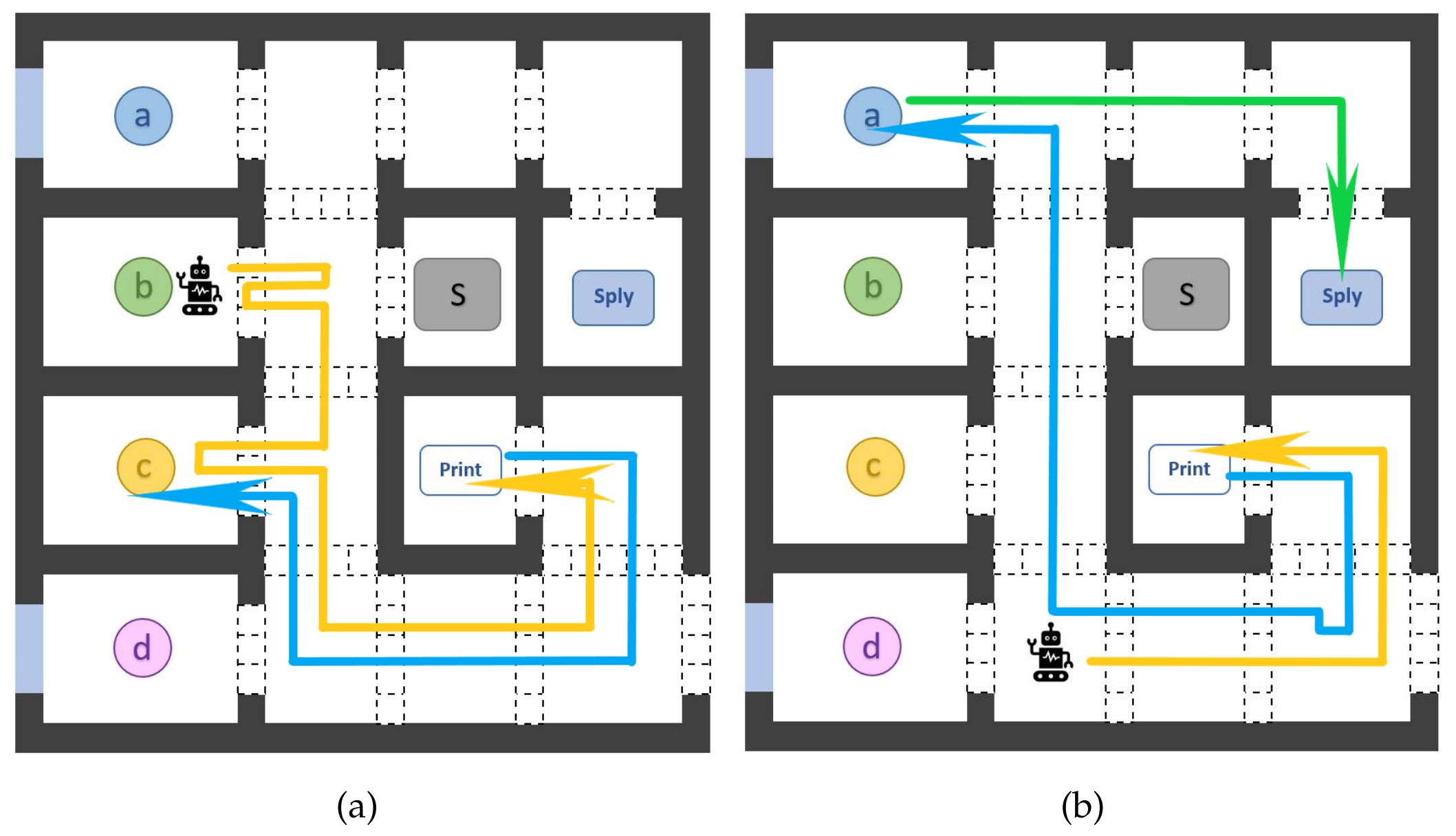}
\caption{Generated paths for the TurtleBot to accomplish tasks if it can observe surroundings in all four directions: (\textbf{a}) Task 1. (\textbf{b}) Task 2.}\centering\label{fig:task_full}
\end{figure}

After the training process is converged, the optimal policy can be derived from Q networks. Figure~\ref{fig:task_full}(a) illustrates a generated path with which the TurtleBot can complete Task 1. It can be seen that after leaving the initial state, office `b', the TurtleBot moves towards the printer room, indicated via a yellow path. It visits offices `b' and `c' more than once on the way to the printer room due to motion uncertainty. After the TurtleBot arrives at the printer room and collects the documents, a blue path demonstrates that it leaves the printer room and moves to office `c' for the delivery.

\paragraph{Task 2:}
Here, we extend Task 1 to a more complex task. Task 2 indicates that the TurtleBot must go to the supply station for recharge after delivering the documents and before repeating Task 1. The LTL formula of Task 2 can be expressed as equation~(\ref{eq:office_task2}), and Figure~\ref{fig:office_task_automaton}(b) depicts the corresponding LDGBA, keeping the transitions with the single label only to simplify the illustration.

\begin{equation}\label{eq:office_task2}
\varphi_{task2}= ( \neg(a \mid c) \mathcal{U} Print ) \wedge ( \neg Sply \mathcal{U} (a \mid c) ) \wedge ( \diamondsuit Sply) \wedge (\square \neg S)
\end{equation}

Figure~\ref{fig:task_full}(b) shows a path generated from the learned policy for the agent to accomplish Task 2. The TurtleBot starts outside office `d'. It then moves to the printer room, collects the documents, and leaves for office `a', indicated as the yellow and blue routes, respectively. Finally, the green route shows that the agent arrives at the `Sply' station for recharge after delivering the documents.

\subsubsection{Observation of a single direction} \label{sec4bb}
\begin{figure}[h]%
\centering
 \includegraphics[scale=0.3]{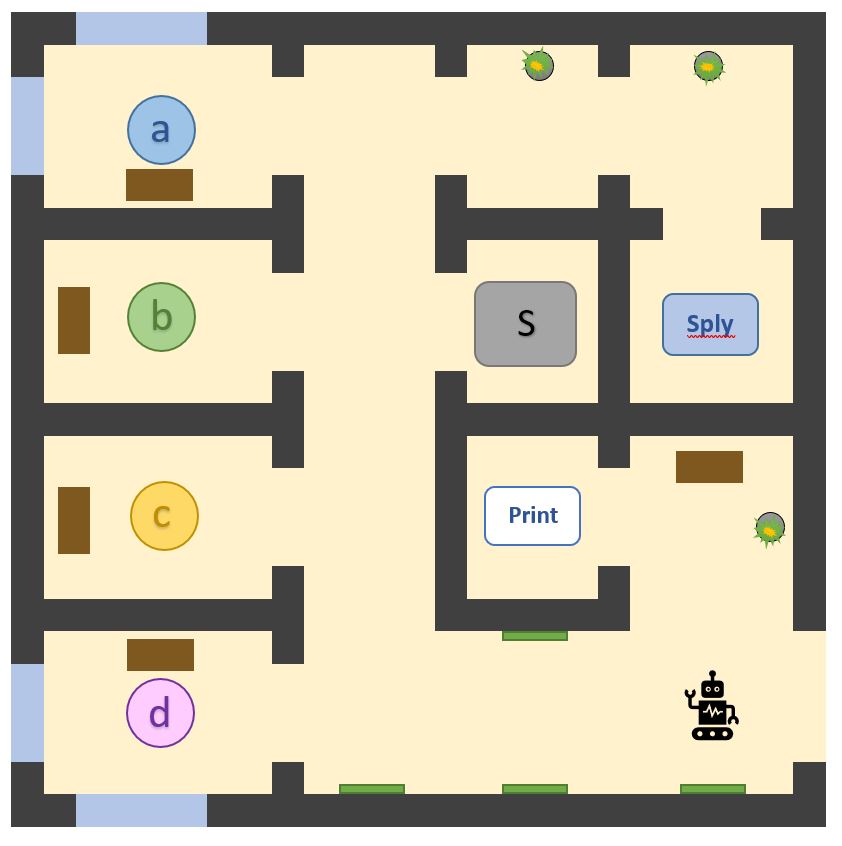}
\caption{The modified office environment where the agent can observe only one direction. }\label{fig:modified_office}
\end{figure}

We also consider another observation setting, in which the TurtleBot is supposed to observe only one direction randomly at each state. Consequently, the observation uncertainty increases significantly. We add a few items in the PyBullet office environment to enhance policy convergence, as shown in Figure~\ref{fig:modified_office}. Therefore, the observable space includes `hallway', `wall', `door', `window', `table', `paint on the wall', and `flower by the wall.’ For example, the agent can observe `wall’, `table’, and `door’ with the probabilities of 50\%, 25\%, and 25\%, respectively, in offices `b’ and `c’.

\begin{figure}[h]%
\centering
\includegraphics[width=100mm]{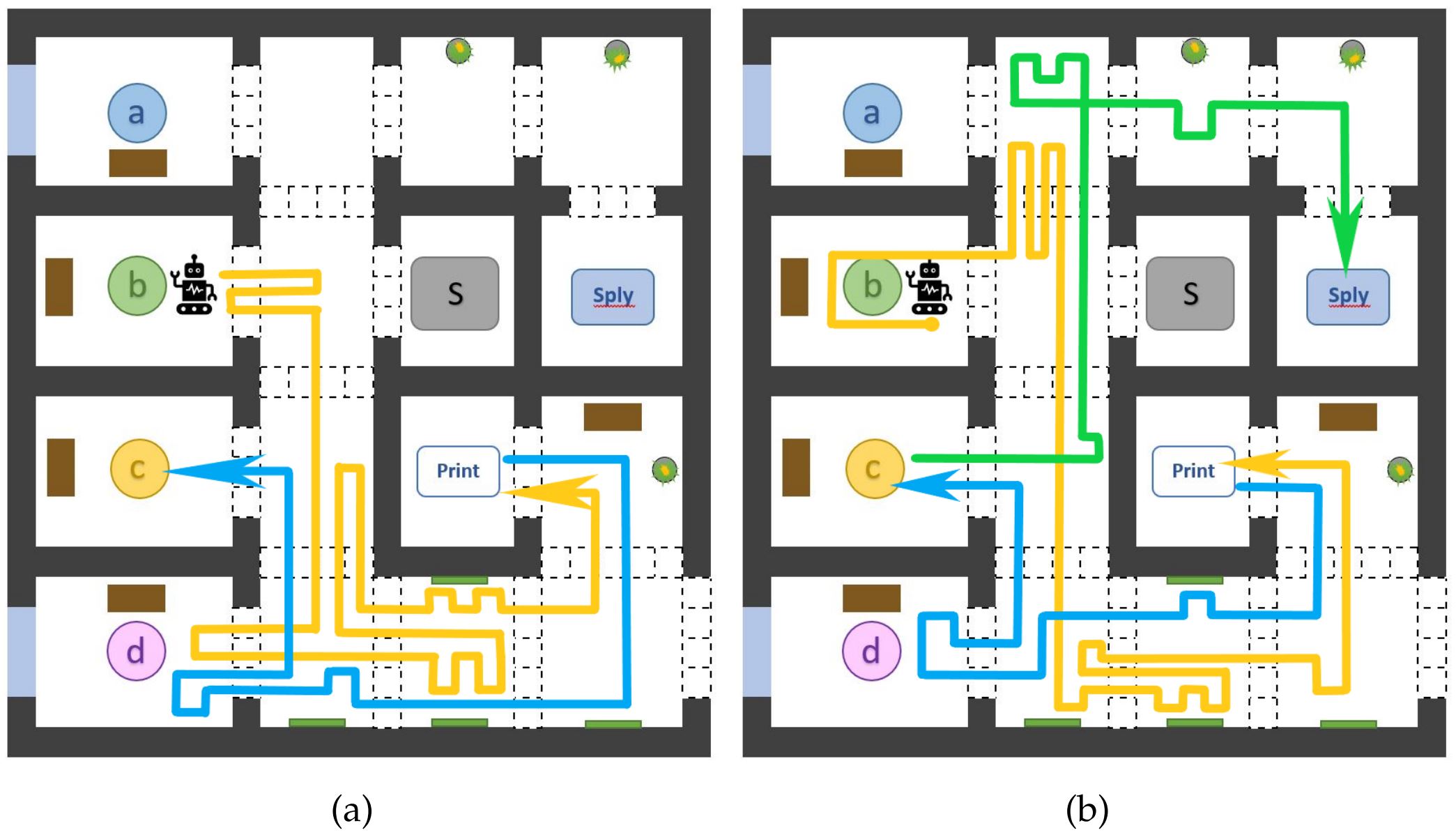}
\caption{The paths for the agent to accomplish the assigned tasks for a single round when the agent can observe only one direction: (\textbf{a}) Task 1. (\textbf{b}) Task 2.}\centering\label{fig:task_single}
\end{figure}

Figure~\ref{fig:task_single} demonstrates the generated paths for the agent to accomplish the same tasks as in Section~\ref{sec5ba}, respectively. Compared with the paths in Figure~\ref{fig:task_full} with four directional observations, the single observation element provides an agent with less sense of the current state. Figure~\ref{fig:task_single} also indicates that the agent encounters difficulty deciding the right moves. For example, the yellow path in Figure~\ref{fig:task_single} (a) illustrates that the agent moves back and forth a few times in the hallway with multiple paints on the wall before finally heading to the printer room. It is mainly because of observation uncertainty in addition to motion uncertainty.

\subsection{Multi-agent Warehouse Simulation}\label{sec5c}

\begin{figure}
\centering
\resizebox*{5cm}{!}{\includegraphics{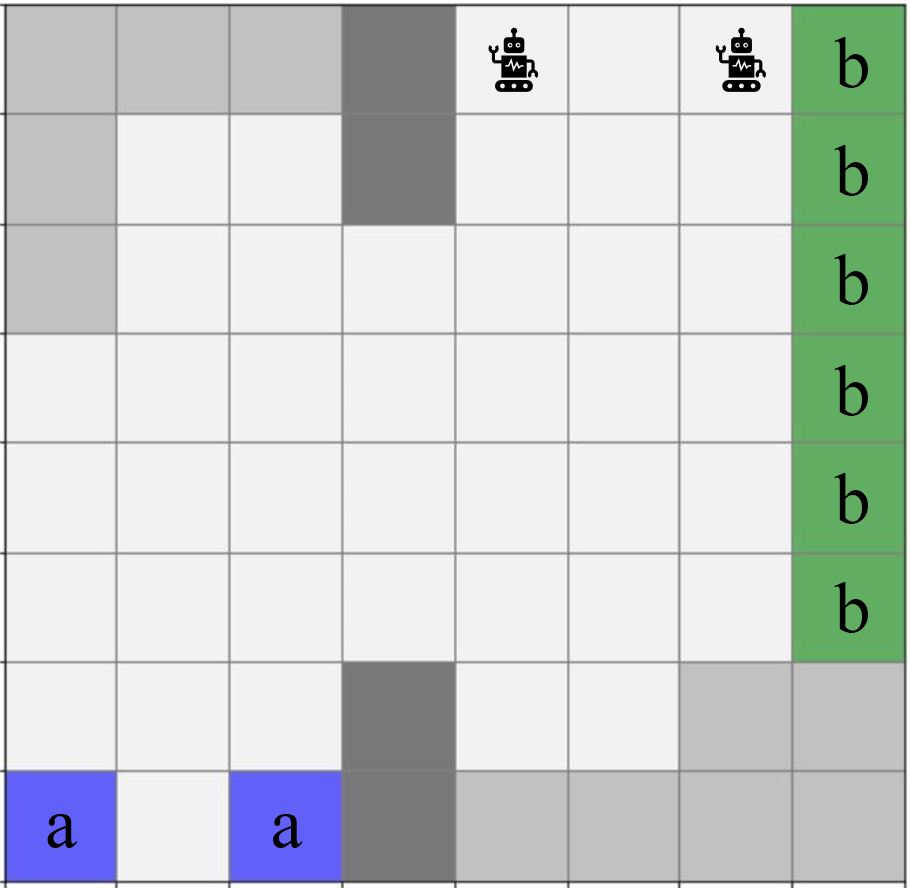}}
\caption{The warehouse environment.}\centering\label{fig:warehouse}
\end{figure}

We also preliminarily apply the developed model-free RL approaches to a multi-agent problem. A mini-factory warehouse is modeled as an $8 \times 8$ grid world as shown in Figure~ \ref{fig:warehouse}. There are two agents, and each must repeatedly move a box from one of the locations labeled as `a' in blue to any spot labeled as `b' in green to drop off the box on the convey belts. The darker and lighter gray states indicate the wall and other immovable packages. In this case, each agent can move `\textit{up}', `\textit{down}', `\textit{right}', `\textit{left}' and `\textit{stay}', but they cannot move to the same location simultaneously. Once both agents complete the task within one round, more boxes await at locations ‘a’. Each agent is rewarded for successfully moving a box to the goal location. In addition, the agents' initial locations are shown in Figure~\ref{fig:warehouse}. The observation and transition probabilities are set as the same as in Section~\ref{sec2c}.

We can formulate the team task via LTL as $\phi = \square \diamondsuit (a \wedge \diamondsuit b)$, similar to the formula in equation~(\ref{eq:gridworld_task2}). Also, we implement the Reward Redesign mentioned in Section~\ref{sec5ab} for task feasibility. Multi-agent reinforcement learning (MARL) studies how multiple agents interact in a common environment. According to agents' actual task requirements, MARL can be categorized as the following three broad classes \cite{Oroojlooy2022}:
\begin{itemize}
    \item Cooperative: Agents cooperate to achieve the team goal, in which no agent can perform the whole task alone.
    \item Competitive: Agents compete against each other.
    \item Mixed: Agents maximize the utility that may require cooperating and/or competing.
\end{itemize}
In addition, the multi-agent system can also be modeled in a centralized or decentralized framework, where a central policy or multiple independent policies can be learned by the agents \cite{Oroojlooy2022}, respectively.

In this case, the agents work in the same environment towards a common goal. This MARL problem can be categorized as a cooperative case. We model this problem as a simplified decentralized POMDP (Dec-POMDP) framework. Each agent has a set of actions $A^i$ and a set of observations $O^i$, where $i$ denotes the index of agents: $i \in \{1,2\}$. However, since they are identical agents in the same state space, we define $A^1=A^2$ and $O^1=O^2$. Only static events are considered in this case. 

\begin{figure}
\centering
\resizebox*{12cm}{!}{\includegraphics{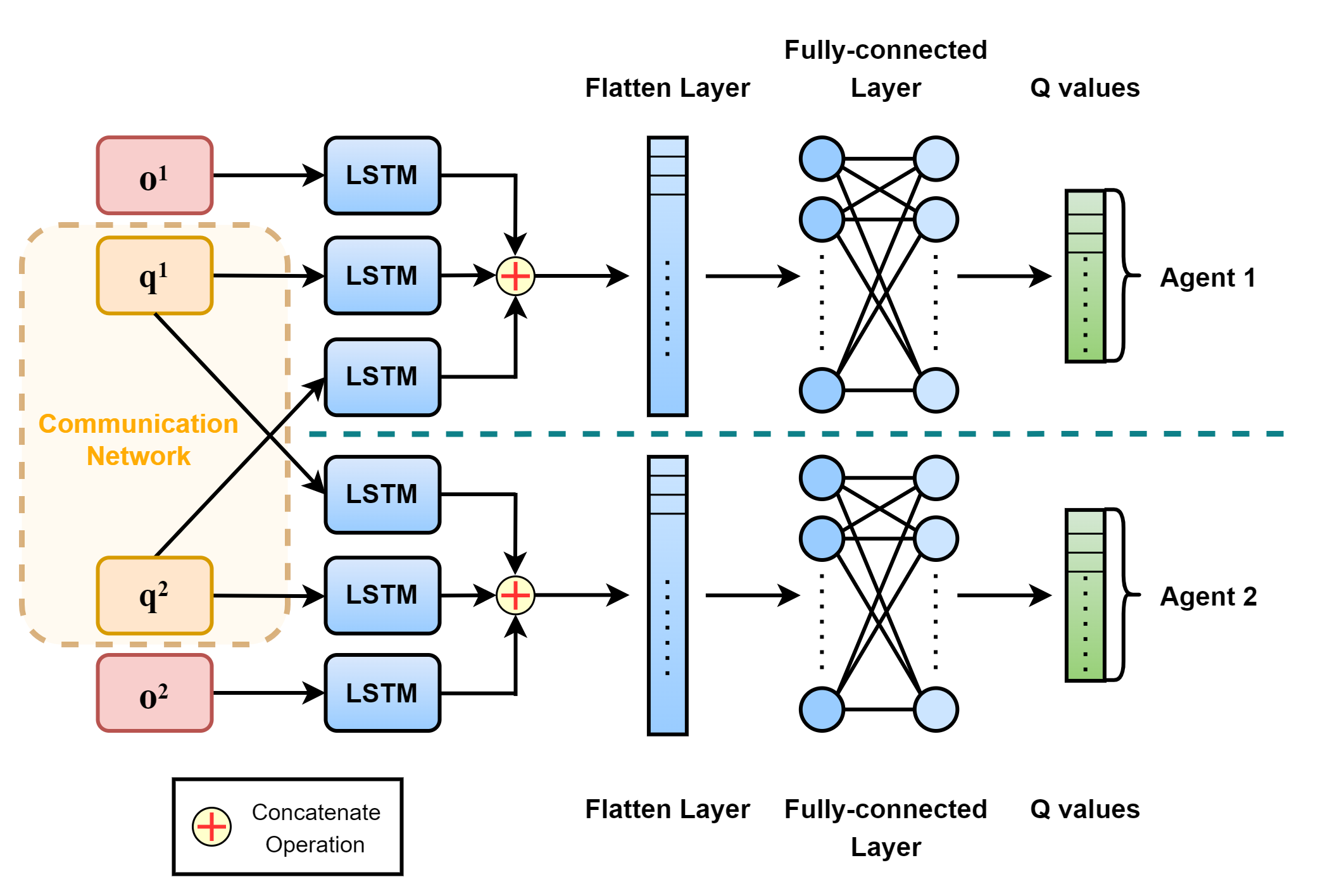}}\centering
\caption{Q network architectures for a MARL problem.}\centering \label{fig:DQN_RNN_MARL}
\end{figure}

\begin{figure}
\centering
\resizebox*{8cm}{!}{\includegraphics{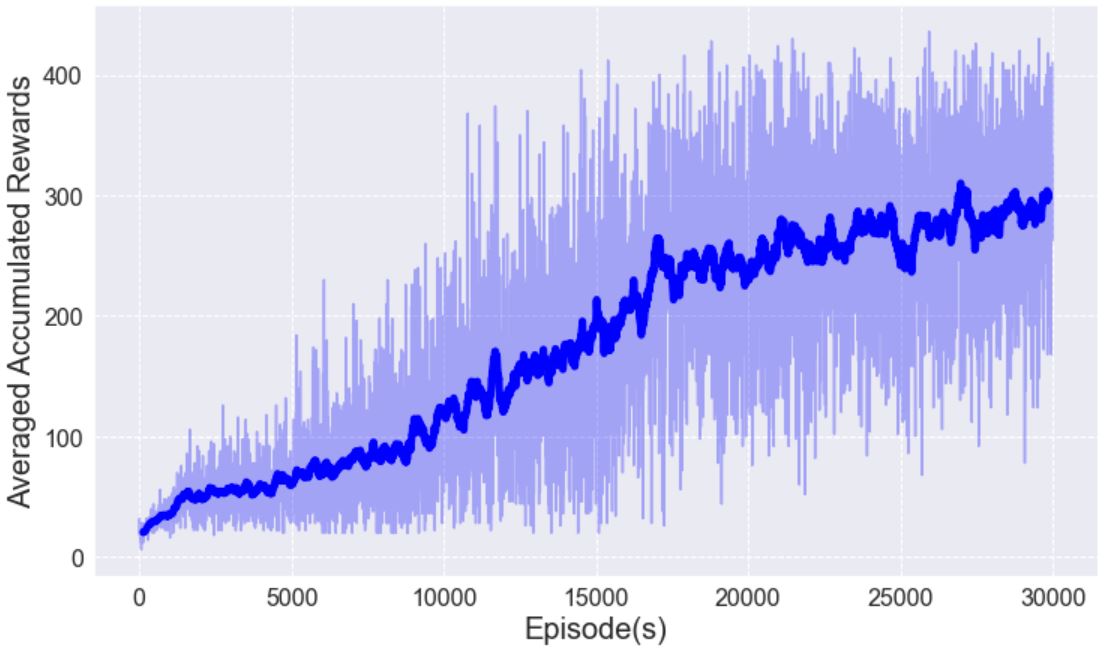}}\centering
\caption{The accumulated rewards of the MARL case.} \label{fig:MARL_plot}
\end{figure}

There have been numerous previous works to solve MARL problems from various perspectives. Inspired by the work of Zhou \textit{et al.} \cite{zhou2022}, we set up cooperative communication between the agents and implement it into our model-free RL algorithms. We assume the agents have full knowledge of the task. Two independent Q networks are initialized for each agent. For each Q network, the input consists of the observation sequence, the $q$ state sequence, and the $q$ state sequence from the other agent. The agents share the task recognition information through the messages between the agents in the same communication network. The corresponding Q networks are shown as Figure~\ref{fig:DQN_RNN_MARL}. Since two Q networks are trained independently to estimate Q values for each agent, we consider this approach a decentralized training and execution. The simulation took 30,000 episodes for 300 steps each. Figure~\ref{fig:MARL_plot} shows the trend of the accumulated reward \textit{vs.} episode.

\begin{figure}[h]%
\centering
\includegraphics[width=130mm]{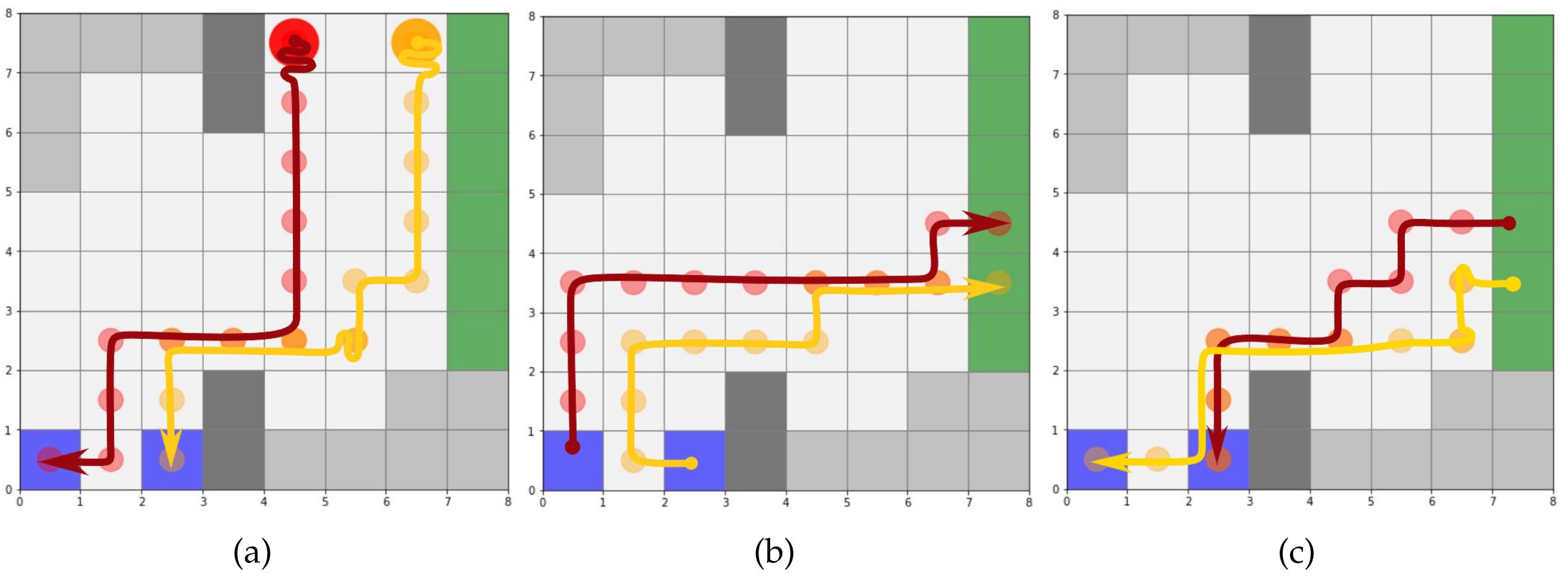}\centering
\caption{The paths generated for the red and yellow agents to accomplish the assigned task.}
\label{fig:MARL_case_path}
\end{figure}

Figure~\ref{fig:MARL_case_path} shows the derived paths for both agents, represented by red and yellow routes, respectively. At the beginning of the task in Figure~\ref{fig:MARL_case_path} (a), both agents stay at their original positions to gather sufficient observations of the surrounding environment for decision-making. Shortly after, they leave for the loading zone in blue, and the yellow agent waits at a state on its way to prevent a collision with the other agent (red). Once the packages are loaded, the agents head towards the green zone to drop them off on the convey belts, as shown in Figure~\ref{fig:MARL_case_path} (b). The second run starts right after the completion, shown in Figure~\ref{fig:MARL_case_path} (c). It shall be noted that the yellow agent selects the action `\textit{stay}' a few times to wait for the other agent to pass because we prioritize the red agent. The generated paths show that the desired cooperative task is achieved even though separate policies are trained for the agents with limited communication.

\section{Discussion}

This study formulates the motion planning of autonomous agents in partially observable environments as a PL-POMDP problem. We also address high-level complex tasks by expressing them through LTLs and converting them to LDGBAs for model checking. Consequently, such a motion planning problem became equivalent to finding an optimal policy on the product of PL-POMDP and the induced LDGBA, and a model-free RL approach is proposed. We employ LDGBAs for model checking because they result in a smaller automaton state space than the corresponding DRAs. In addition, they have multiple accepting sets, enabling the agent to visit all accepting sets infinitely often. With the implementation of the reward redesign, the modified LDGBA can address the sparsity of rewards caused by LDBA in RL for motion planning \cite{Oura2020}. In the future, we plan to leverage the tracking-frontier function proposed in \cite{Cai2021b} to keep track of non-visited accepting sets in LDGBA. This could be particularly useful when more complicated surveillance tasks for UAVs are required in partially observable environments. By further developing this function, we aim to improve the performance and applicability of our proposed method. 

Deep Q learning is employed in the proposed model-free RL approach to learning optimal policies on the product PL-POMDP. Specifically, LSTM is implemented into Q network architectures to process the agent's observation history and task recognition. Using the induced automation state or perceived label history to represent task recognition depends on whether the agent is fully aware of the task. As the simulation examples demonstrate, either representation can enhance the agent’s learning in partially observable environments subject to complex tasks. We also investigate the performances of Q networks by using different lengths of the input sequences. Choosing too long or too short sequences can lead to an unstable training process. The sequence lengths used in this study provide the agent with sufficient information for decision-making and avoid lengthy delays before initiating action selection and following the trained policy. It shall be noted that the length selection may depend on the solved problem. The proposed approach can be easily updated with other value-based or policy-based deep reinforcement learning (DRL) methods in future work. 

We apply the proposed model-free RL approach to the motion planning of autonomous agents in discrete environments only in this study. Future works will extend the approach to POMDP problems with continuous state and action spaces. In addition, a general LDGBA can consist of non-deterministic and deterministic sets. Although most LDGBA corresponding to the tasks considered in our simulation examples have deterministic sets only, one example demonstrates that the proposed approach can handle $\epsilon$-transitions, which are non-deterministic. Even if the initial state is in the non-deterministic set, the automaton state transitions are restricted in the deterministic set after an $\epsilon$-transition. Because the accepting states are in the deterministic set only, the proposed approaches can still achieve optimal policies. 

We also provide a preliminary simulation of a cooperative multi-agent system to demonstrate the proposed approach's versatility and adaptability. Furthermore, by enabling the communication of task recognition between the agents, the acquired policies show a high level of collaboration that results in successful task completion. Although the current approach in the multi-agent simulation focuses on a simplified scenario by using a decentralized training and execution framework, which can often be non-stationary \cite{Oroojlooy2022}, it has the potential to address a system of multiple UAVs for cooperative missions. In the future, it may be more appropriate to use a centralized training and decentralized execution (CTDE) framework capable of handling a continuous state space. Moreover, the agents' task recognition communication relies on their full knowledge of the tasks. An alternative approach can use local observations from other adjacent agents, so an additional neural network is needed to approximate the communication mechanism.

\backmatter





\bmhead{Acknowledgments}

Li and Xiao would like to thank US Department of Education (ED\#P116S210005) and NSF (\#2226936) for supporting this research.

\section*{Declarations}

\begin{itemize}
\item Funding: This research was funded by US Department of Education (ED\#P116S210005) and NSF (\#2226936).
\item Competing interests: The authors declare no competing interests.

\item Code availability: Code files are available at \url{https://github.com/JunchaoLi001/Model-free_DRL_LSTM_on_POMDP_with_LDGBA}
\end{itemize}



\bibliography{sn-article}

\end{document}